\documentclass[10pt,twocolumn,letterpaper]{article}

\usepackage{wacv}
\usepackage{times}
\usepackage{epsfig}
\usepackage{graphicx}
\usepackage{amsmath}
\usepackage{amssymb}

\usepackage{xcolor,colortbl}
\usepackage{subfigure}
\usepackage{url}
\usepackage[stable]{footmisc}
\usepackage{subfigure}

\def\wacvPaperID{9} 

\wacvfinalcopy 

\ifwacvfinal
\fi

\ifwacvfinal
\usepackage[breaklinks=true,bookmarks=false]{hyperref}
\else
\usepackage[pagebackref=true,breaklinks=true,colorlinks,bookmarks=false]{hyperref}
\fi

\ifwacvfinal
\else
\fi

\begin{document}

\pagenumbering{gobble} 

\title{Person Perception Biases Exposed: Revisiting the First Impressions Dataset}


\author{Julio C. S. Jacques Junior\\
Universitat Oberta de Catalunya, Spain\\
Computer Vision Center, Spain\\
{\tt\small jsilveira@uoc.edu}
\and
Agata Lapedriza\\
Universitat Oberta de Catalunya, Spain\\
{\tt\small alapedriza@uoc.edu}
\and
Cristina Palmero\\
Universitat de Barcelona, Spain\\
Computer Vision Center, Spain\\
{\tt\small crpalmec7@alumnes.ub.edu}
\and
Xavier Bar\'o\\
Universitat Oberta de Catalunya, Spain\\
Computer Vision Center, Spain\\
{\tt\small xbaro@uoc.edu}
\and
Sergio Escalera\\
Universitat de Barcelona, Spain\\
Computer Vision Center, Spain\\
{\tt\small  sergio@maia.ub.es}
}





\maketitle

\begin{abstract}
This work revisits the ChaLearn First Impressions database, annotated for personality perception using pairwise comparisons via crowdsourcing. We analyse for the first time the original pairwise annotations, and reveal existing person perception biases associated to perceived attributes like gender, ethnicity, age and face attractiveness. We show how person perception bias can influence data labelling of a subjective task, which has received little attention from the computer vision and machine learning communities by now. We further show that the mechanism used to convert pairwise annotations to continuous values may magnify the biases if no special treatment is considered. The findings of this study are relevant for the computer vision community that is still creating new datasets on subjective tasks, and using them for practical applications, ignoring these perceptual biases. 
\end{abstract}

\section{Introduction}
Psychologists have long studied human personality, and throughout the years different theories have been proposed to categorise, explain and understand it. From the past few years, it has also become an attractive research area in visual computing~\cite{Vinciarelli:2013,jacques:TAC:2019}, motivated by the fact that automatic methods for personality recognition or perception can be applied in a vast number of scenarios. Nevertheless, while real personality can be accessed through self-report questionnaires, perceived (or apparent) personality assessment is given by external observers through impression formation, and here is where person perception bias comes in. 

Technologies for human behaviour analysis have shown their vulnerability to human annotation biases~\cite{Escalante:TAC:2020}. In particular, human bias is very strong when trying to infer personality attributes of someone during a first short encounter. This subjectivity makes the task of creating automatic personality perception systems challenging, since the biases will be reflected on the annotations and, consequently, on the resulting recognition systems. Therefore, creating methods that preserve human bias can have negative consequences if they are used in applications that deal with human outcomes. 
While the use of pairwise instance comparison~\cite{lopez2016chalearn,Chen2016,Joo:ICCV:2015} significantly reduces perception bias produced by absolute annotations, completely eliminating it in subjective tasks is extremely difficult. 

\begin{figure*}[htbp] 
\centering
  \includegraphics[width=1.0\linewidth]{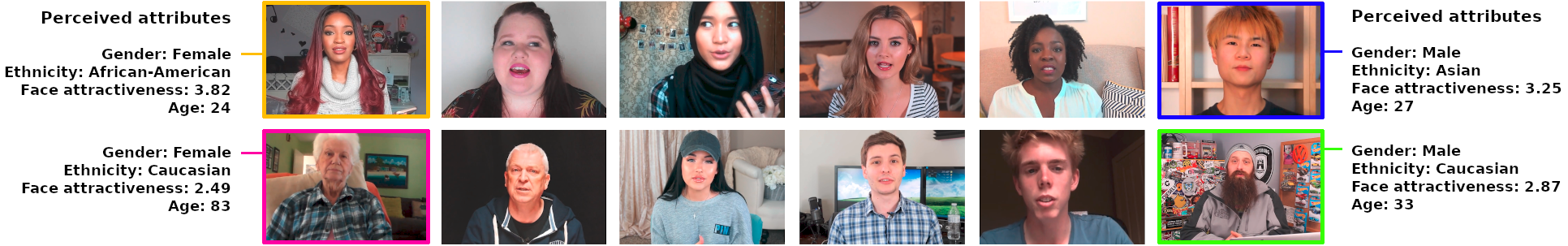}
  \caption{Imagine that pairs of short videos are given, with people talking to the camera about any predefined topic. As annotator, you are asked to define what individual in a pair looks more friendly, more organised or maybe more authentic. Then, you may start analysing people's behaviour and attributes in order to build your first impressions. At the end, your choices might tell something about you. However, the overall perception given hundreds or thousands annotators may tell something about the database. Snapshots from the First Impressions~\cite{lopez2016chalearn} dataset (attributes empirically defined for illustration purposes).}
  \label{fig:teaser}
\end{figure*}

This work uses the First Impressions (FI)~\cite{lopez2016chalearn}  dataset to expose the existence of person perception bias in data labelling of personality. The FI dataset is one of the biggest publicly available datasets on the topic. Our work is based on recent studies that demonstrate the bias produced by perceived gender, attractiveness and age~\cite{dario:TAC:2019} during the impression formation. In particular, we derive perception biases from pairwise annotations and associated person's attributes\footnote{Attribute categories used in this research are imperfect for many reasons. For example, there is no gold standard for ``ethnicity'' categories, and it is unclear how many gender categories should be stipulated (or whether they should be treated as discrete categories at all). This work is based on an ethical and legal setting, and the methodology and findings are expected to be applied later to any re-defined and/or extended attribute category.}. For example, we show that women are more frequently perceived as more \textit{Open to experience} than men, that older men are more frequently perceived as more \textit{Conscientious} than younger ones, and that ethnicity has stronger influence than gender if African-Americans are compared to either Asians or Caucasians, which bring to light some annotators' bias.  Fig.~\ref{fig:teaser} illustrates how person perception can influence data labelling when a subjective task like personality perception is considered. 

Supervised learning methods developed to recognise apparent personality from images or videos~\cite{jacques:TAC:2019} require a label for each individual in the train data, and pairwise annotations are in general not used. For this reason, the pairwise annotations of the FI~\cite{lopez2016chalearn} dataset were originally converted to continuous values using~\cite{Chen2016}. Our study also reveals that the mechanism used to convert pairwise annotations to continuous values may magnify the biases, making stereotyping stronger. 
Finally, it is important to note that previous works (e.g.,~\cite{jacques:TAC:2019,Escalante:TAC:2020,Yan:ICMI:2020}) using the FI dataset are based on the continuous values originated from the pairwise labels, and this is the first time the original (raw) pairwise labels are analysed.


\section{Ethical Implications\footnote{For more information about ethics in AI, we refer the reader to~\cite{ethics:ai:eu}.}}\label{sec:ethics}

\textbf{Personality perception and its applications}. People spontaneously build first impressions of unacquainted individuals in milliseconds, even from a still photograph, quite consistently~\cite{Todorov:2017:book}. However, such snap judgements, which are built and used to interact with others, are often stereotyped~\cite{Riva:2019}. Therefore, do we want machines to do the same? Having machines that form first impressions of others has risks. Such systems are trained from human annotations and inherit human perception bias along with other biases created by culture, beliefs or previous experiences. Since it is highly likely that automatic personality perception is not accurate, these technologies are not ready to be used for legal applications or for anything that determines opportunities for people, such as job interviews. Furthermore, having access to people's personality (either if real or apparent) just by extracting and analysing data from any kind of input could represent a major threat to their privacy. Not only in terms of rights, but also because it could pave the way for effective mass manipulation and psychological persuasion~\cite{Matz:2017}. On the other side, automatic personality perception can be very useful in social robotics~\cite{breazeal2004social}, to design machines that can approach people in a natural way, creating more comfortable experiences and building trust~\cite{LANGER2019231}. In particular, applications related to health care, education or human assistance can benefit from using automatic impression formation.

\textbf{Bias in face attributes recognition}. 
Our work partially relies on automatic face classification methods to extract an attractiveness score and to estimate apparent age. Both methods suffer from the same type of perception bias previously described. According to~\cite{Talamas:2016}, facial cues often guide first impressions and these first impressions guide our decisions. Face attractiveness, however, is very subjective and may be subject to critics when applied to social computing. Nevertheless, the topic has been widely studied in psychology/sociology~\cite{Timmerman:1980,Lucker:1981,Zebrowitz:2014,Palmer:2016,Talamas:2016}. These attributes have been selected to give visibility to the existing biases, especially because the well known ``\textit{attractiveness halo effect}'' (i.e., more positive impressions are given to more attractive people) has its particular influence on data labelling. 



\section{Related Work}
Fairness in machine learning~\cite{Bird:2019:FML} is rapidly gaining interest among the research community and industry. This has been partially motivated by the biased results reported in the literature (e.g.,~\cite{Zhao:2017,Hendricks_2018_ECCV,Escalante:TAC:2020,dario:TAC:2019}), along with the difficulty to interpret \textit{latent} representations~\cite{Quadrianto_2019_CVPR}. According to \cite{friedler2019comparative}, fairness-aware machine learning approaches can be categorised as: 1) preprocessing techniques which aim to modify the input data; 2) algorithm modification techniques, which modify existing algorithms by adding constraints or regularisation; and 3) postprocessing techniques which modify the output of an existing method to be fair. These categories consider the data is already available and ready to use. Our work, however, goes one step back and analyses how perception bias affects data labelling of a subjective task, which aligns with the idea that unfairness induced by 
unmeasured predictive variables should be addressed through data collection~\cite{chen2018my}. 
Thus, rather than addressing general bias problems such as imbalanced training data, covariate shift or sample selection~\cite{gu2019understanding}, which can be found in almost any machine learning-based task, this work focuses on the biases coming from \textit{human perception}.

In visual perception, contextual effects and prior experience lead to systematic biases in the judgement~\cite{Dekel:2020}. Cognitive and perceptual biases have distinct causes and effects, and can be grouped into different categories~\cite{Dimara:2020} given the bias type (e.g., fundamental attribution error, cultural bias, belief bias, selective perception, among others). The biases produced by human perception, which have been widely studied in sociology and psychology (e.g.~\cite{todorov:2019:gender,Talamas:2016}), have a strong influence in subjective tasks such as automatic personality perception~\cite{jacques:TAC:2019}, (job) recommendation systems~\cite{Escalante:TAC:2020}, emotion recognition~\cite{Shen:2019:ACII} or image captioning~\cite{Hendricks_2018_ECCV}. However, works from a psychological perspective are limited to perform statistical analysis on small-scale datasets. On the other hand, most works from a computational perspective~\cite{Torralba:cvpr:2011,pmlr-v81-buolamwini18a,Zhao:2017,Jiang:2019,Wang_2019_ICCV,Yucer_2020_CVPR_Workshops,Sixta:eccvw:2020} study the general bias problems~\cite{gu2019understanding} mentioned above, while little attention is given to subjective bias analysis~\cite{Shen:2019:ACII,Quadrianto_2019_CVPR,Robinson_2020_CVPR_Workshops,Yan:ICMI:2020} beyond the perspective of explainable models~\cite{Escalante:IJCNN:2017,Park:CVPR:2018,dario:TAC:2019,Escalante:TAC:2020}.

In~\cite{Shen:2019:ACII}, authors show that the order of how images are displayed to the annotators may significantly bias the labels in facial emotion recognition tasks, whereas~\cite{Quadrianto_2019_CVPR} proposed a data-to-data translation approach by learning a mapping from an input domain to a fair target domain, where a fairness constraint is enforced. The latter focused on analysing the gender attribute and the overall goal was to maximise equal opportunity between males and females. Robinson et al.~\cite{Robinson_2020_CVPR_Workshops} showed that the performance gaps in face recognition can be reduced by learning subgroup-specific thresholds, revealing that the conventional approach of learning a global threshold may also bias the results. More recently, Yan et al.~\cite{Yan:ICMI:2020} investigated the biases on multimodal systems designed for automatic personality perception, using the FI~\cite{lopez2016chalearn} dataset as case study. The study revealed that different modalities show various patterns of biases, and that data fusion also introduces additional biases to the model. Thus, they propose two debiasing approaches based on data balancing and adversarial learning to mitigate the biases. The analyses performed in their work, however, are based on the continuous values provided with the FI~\cite{lopez2016chalearn} dataset, and the original pairwise annotations are not considered.

Collecting labels for subjective tasks is challenging. Biased annotations are particularly difficult to detect and correct. 
For annotation tasks related to subjective human behaviour and personality attributes~\cite{jacques:TAC:2019}, pairwise comparison is becoming a standard procedure, as it has demonstrated~\cite{lopez2016chalearn,Chen2016,Joo:ICCV:2015} to be very effective at mitigating labeller biases. For instance, Joo et al.~\cite{Joo:ICCV:2015} asked Amazon Mechanical Turk workers to compare a pair of images in face trait dimensions rather than evaluating each image individually. A similar strategy was applied in~\cite{lopez2016chalearn,Escalante:IJCNN:2017,Escalante:TAC:2020} for video files.

Comparison schemes have three main advantages in data labelling for person perception: 1) they naturally identify the strength of each sample in the context of relational distance from other examples, generating a more reliable ranking of subtle signal differences~\cite{Joo:ICCV:2015}; 2) they mitigate the sequential bias~\cite{Shen:2019:ACII}, e.g., scoring someone very low on a certain dimension because of an unconscious comparison with previous samples where the score was high; and 3) the annotators do not need to establish the absolute baseline or scales for any dimension, which would be unnatural. Although pairwise ratings significantly reduce the bias in person perception annotation tasks, this work shows that people's attributes, combined with annotators' bias, can have a strong influence on data labelling. This suggests that future works on the topic need to pay attention to the way the pairs are defined and presented to the annotators, since the pairs themselves can also be a source of bias, particularly for sensitive applications where reducing biases under certain controlled dimensions is crucial.


\section{The First Impressions Dataset}\label{sec:FIdataset}

\begin{figure}[htbp] 
	\centering    
    \includegraphics[height=5.0cm]{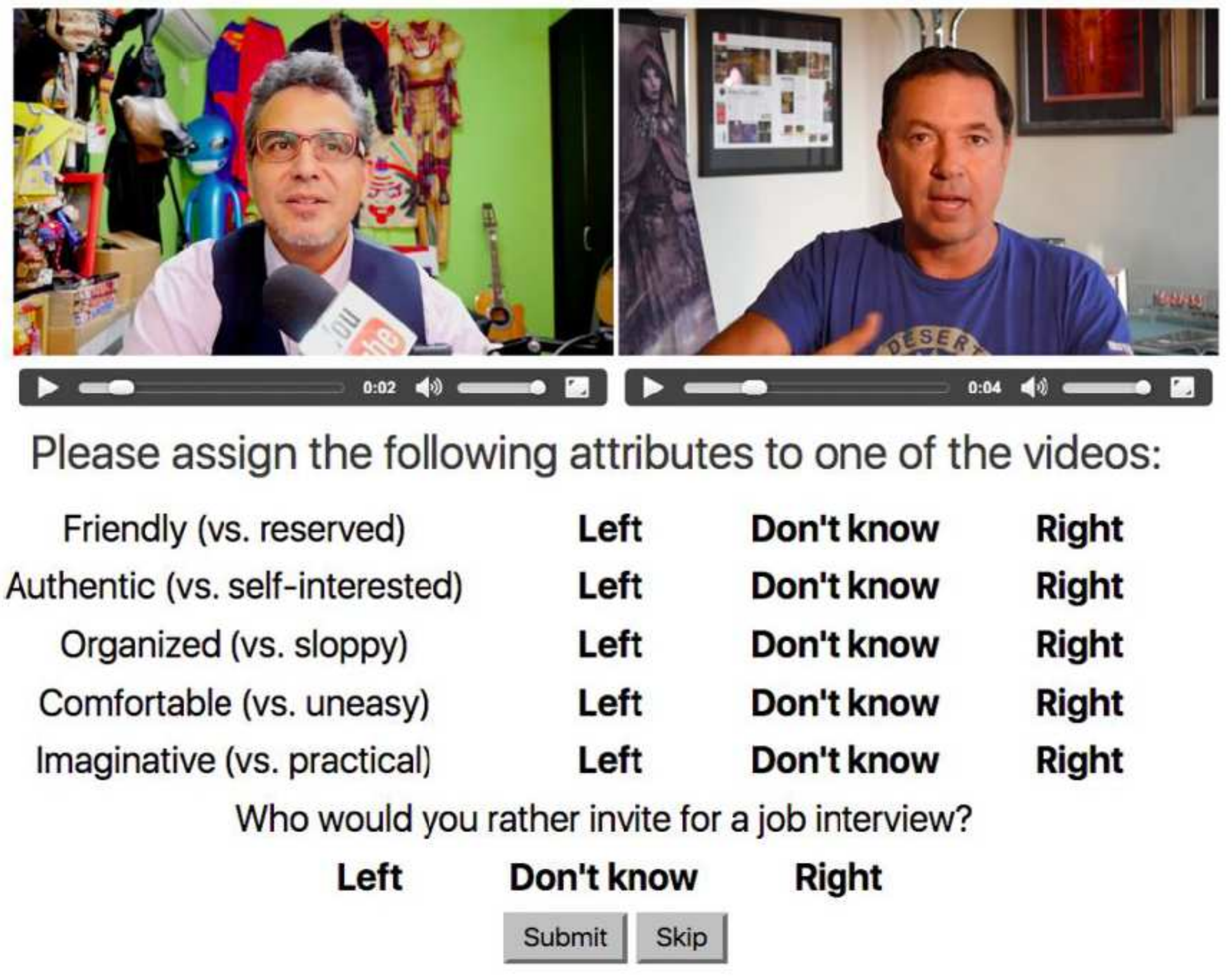}
	\caption{The interface used for pairwise data labelling~\cite{lopez2016chalearn}.}
	\label{fig:interface}
\end{figure}

The ChaLearn First Impressions (FI) dataset~\cite{lopez2016chalearn} is currently the largest, public and labelled dataset developed to advance research on automatic personality perception. 
The FI dataset was released in the context of a computational challenge, where the goal was to automatically recognise the Big-Five (OCEAN) apparent personality traits of single individuals in videos:  \textit{\textbf{O}penness to experience}, \textit{\textbf{C}onscientiousness}, \textit{\textbf{E}xtraversion}, \textit{\textbf{A}greeableness}, and \textit{\textbf{N}euroticism}\footnote{\textit{Neuroticism} was labelled in~\cite{lopez2016chalearn} as \textit{``Emotion stability''}, which is the opposite of Neuroticism. This will be represented along the paper as $\overline{\mbox{N}}$.}. Later, it was labelled with an ``Invite to interview'' variable, aiming to advance research on explainable machine learning~\cite{Escalante:IJCNN:2017}. The dataset is composed of 10K short video clips (average duration of 15s each) extracted from more than 3K  different YouTube videos of people talking to a camera. Some snapshots of the dataset are shown in Fig.~\ref{fig:teaser}, while Fig.~\ref{fig:interface} shows the pairwise-based annotation interface. The database was annotated using crowdsourcing, being each pair annotated by one single annotator. 
In this work, we release\footnote{The pairwise annotations of the FI dataset can be found at \url{http://chalearnlap.cvc.uab.es/dataset/24/description/}.} and analyse by the first time the original pairwise annotations of the First Impressions dataset.

Gender and ethnicity labels are also provided with the data (both provided via crowdsourcing, i.e., they are perceived attributes). Overall, the dataset is more or less balanced in gender (45\% males and 55\% females). However, it is very imbalanced in terms of ethnicity (3\% Asian, 86\% Caucasian and 11\% African-American). 


The dataset has around 345K video pairs labelled with apparent Big-Five personality traits and the ``Interview'' variable. However, some pairs were labelled with the ``\textit{Don't know}'' label (illustrated in Fig.~\ref{fig:interface}) for some dimensions, for which the annotators were not so confident about the ranking. Table~\ref{tab:validpairs} shows the number of ``valid pairs'' (i.e., when ignoring the ``\textit{Don't know}'' label) per dimension, as well as for different subsets given the gender/ethnicity of individuals being compared. As it can be seen, data imbalance is strong with respect to the different subsets, which imposes another obstacle in addition to perception bias when the goal is to build fair machine learning methods.

\begin{table}[htbp]
\centering
\setlength\tabcolsep{3.0pt}
\scriptsize
\caption{Number of ``valid pairs'' per trait and per subset, given the gender/ethnicity of individuals in a pair.}
\begin{tabular}{|c|r|r|r|r|r|r|}
\hline
\multicolumn{1}{|l|}{} & \multicolumn{1}{c|}{\textbf{O}} & \multicolumn{1}{c|}{\textbf{C}} & \multicolumn{1}{c|}{\textbf{E}} & \multicolumn{1}{c|}{\textbf{A}} & \multicolumn{1}{c|}{\textbf{$\overline{\mbox{N}}$}} & \multicolumn{1}{c|}{\textbf{Interview}} \\ \hline
\textbf{Valid pairs} & 307513 & 313749 & 321684 & 318792 & 321078 & 323178 \\ \hline
\multicolumn{7}{|c|}{\textit{\textbf{Per Gender}}} \\ \hline
\textit{Male vs. Female} & 152365 & 155466 & 159467 & 157829 & 158942 & 160095 \\ \hline
\textit{Female vs. Female} & 91931 & 93795 & 96231 & 95483 & 96080 & 96711 \\ \hline
\textit{Male vs. Male} & 63217 & 64488 & 65986 & 65480 & 66056 & 66372 \\ \hline
\multicolumn{7}{|c|}{\textit{\textbf{Per Ethnicity}}} \\ \hline
\textit{Cauc. vs. Cauc.} & 227558 & 232160 & 238080 & 235944 & 237546 & 239142 \\ \hline
\textit{Afr-Am. vs. Cauc.} & 56496 & 57586 & 59100 & 58543 & 59008 & 59387 \\ \hline
\textit{Asian vs. Cauc.} & 17367 & 17770 & 18137 & 18011 & 18155 & 18264 \\ \hline
\textit{Afr-Am. vs. Afr-Am.} & 3558 & 3642 & 3694 & 3654 & 3710 & 3702 \\ \hline
\textit{Asian vs. Afr.} & 2204 & 2261 & 2330 & 2296 & 2311 & 2341 \\ \hline
\textit{Asian vs. Asian} & 330 & 330 & 343 & 344 & 348 & 342 \\ \hline
\end{tabular}
\label{tab:validpairs}
\end{table}


Having the data labelled through pairwise comparisons, the pairwise data is converted in~\cite{lopez2016chalearn,Escalante:IJCNN:2017} to continuous values using~\cite{Chen2016}. 
This method individually converts the ordinal ratings of each dimension into continuous values (such as the level of ``\textit{Extraversion}'') by fitting a Bradley-Terry-Luce (BTL) model with maximum likelihood, which are further scaled to be in the range of $[0,1]$. This way, each video sample in the dataset will have a continuous value associated to each trait dimension, which can be used by any supervised learning method, in a classification or regression task.

\section{Automatic extraction of face attributes}\label{sec:faceattributeextraction}

This section describes how face attractiveness and perceived age of people present in the FI dataset are obtained. To remove any bias caused by the imbalanced ethnicity category, only Caucasian individuals are considered. 

\begin{figure}[htbp]
	\centering
	\includegraphics[width=1.0\linewidth]{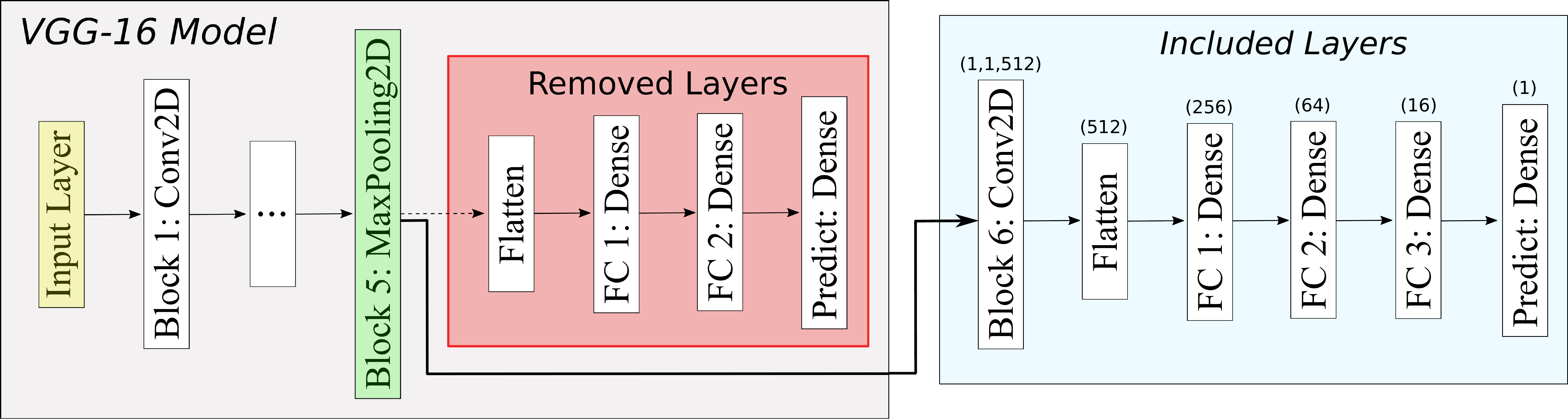}
	\caption{Modified VGG-16 model used to predict either face attractiveness or perceived age (depending on the given task).}
	\label{fig:vgg16-model}
\end{figure}

First, a face detector~\cite{75Zhang:SPL:2016} is applied on each video at $5$ consecutive frames. Then, face attributes are extracted using a modified version of the VGG-16~\cite{simonyan2014deep} model, that regresses either the attractiveness score or the perceived age, depending on the given task. Finally, the per-frame predictions are averaged per attribute. 
The proposed modification consists in removing the last layers of the original VGG-16 model (illustrated in Fig.~\ref{fig:vgg16-model} by a red box) and the inclusion of a convolutional layer (to reduce dimensionality) and three fully connected (FC) layers to learn hidden representations (using ReLu as activation function) before a final Dense layer (with \textit{Sigmoid} activation) responsible for regressing the face attribute. 

\begin{figure}[htbp]
	\centering
	\includegraphics[width=1.0\linewidth]{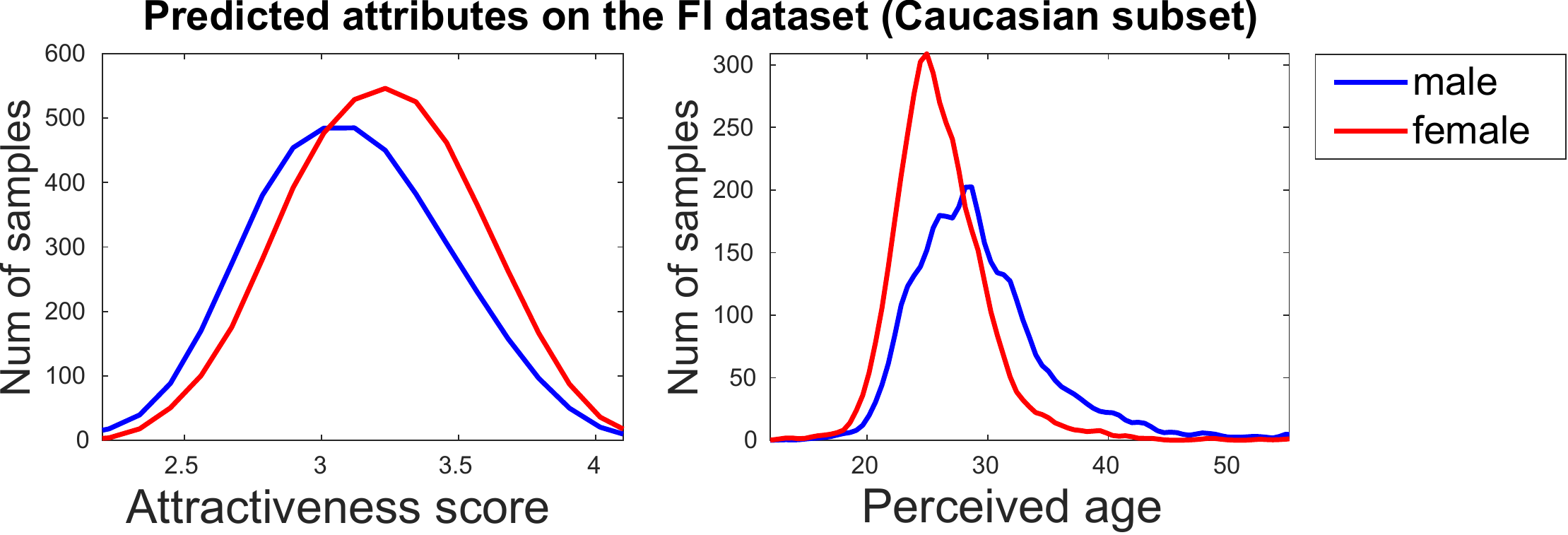}
\caption{Distributions of predicted face attributes.}
\label{fig:att_age_dist}
\end{figure}

Fig.~\ref{fig:att_age_dist} shows the distribution of face attributes extracted for all Caucasian individuals in the FI dataset. It must be emphasised that the aim of our work is not to advance the state of the art on face attribute recognition. Predicted attributes are taken as ``truth'' (i.e., soft labels, more precisely) due to the low error rates obtained on the associated datasets, detailed next, and used as proof of concept. 

\subsection{Face attractiveness}
To recognise the attractiveness score of each individual, our model was trained with the SCUT-5500~\cite{liang2018scut} database. This dataset consists of 5.5K frontal unoccluded faces, with neutral expression, aged from 15 to 60 years old. It contains 4K images of Asians and 1.5K images of Caucasians, equally distributed in gender for each set. Images were labelled with beauty scores in the range of $[1,5]$ by a total of 60 volunteers aged from 18-27, which is also subject to impact the ground truth due to their implicit bias~\cite{Zebrowitz:2014}. 

To evaluate the effectiveness of our model to predict attractiveness score on the SCUT-5500~\cite{liang2018scut} database, $85\%$ of the data was randomly selected for training and the remaining samples for testing. Obtained Mean Absolute Error (MAE) on the test set was $0.247$, comparable to results obtained in~\cite{liang2018scut}. Note that we have also evaluated our model following the same protocol described above but considering Caucasian individuals only, however, obtaining a slightly higher MAE, most probably due to difficulty to generalise given the small-sized training data.

\subsection{Perceived age}
To automatically recognise the perceived age of each individual, the APPA-REAL~\cite{Agustsson2017} database was chosen. The database is composed of almost 8K images mainly showing a single person in frontal face, labelled with real and apparent age (in the latter case, via crowdsourcing), ranging from 0 to 95 years old. Our study, however, uses only the perceived age label as our intention is to analyse how the perception of age can bias pairwise data labelling. On average, each image was annotated with apparent age by 38 annotators, resulting in a very stable average apparent age ($0.3$ standard error of the mean).  

To evaluate the performance of our VGG16-based model to predict apparent age, we followed the evaluation protocol defined in~\cite{Agustsson2017}. Obtained MAE on the test set was $7.12$ (years), which is similar to results obtained by~\cite{jacques:FG:2019}.

\subsection{Training strategy}

The two face attribute recognition tasks are trained in two stages. First, the model is initialised with weights pretrained on ImageNet~\cite{Imagenet:2009}. Then, we train only the new layers. In a second stage, we fine-tune the whole model. Adam algorithm is used as optimiser, with learning rate $1\mathrm{e}{-5}$. Mean Squared Error is used as the loss function. Early stopping is performed if no decrease in validation error is observed. Finally, the best model for each task is kept based on the accuracy computed on the validation set.

\section{Revealing the perception biases}\label{sec:exposingbias}
This section reveals different perception biases found in the FI~\cite{lopez2016chalearn} dataset, from a global to a fine-grained analysis. In Sec.~\ref{subsec:genderbias} and Sec.~\ref{subsec:etn:gender}, we analyse perception biases associated to gender and ethnicity obtained directly from the pairwise (binary) labels, and show that some of them are amplified when converted to continuous values using~\cite{Chen2016}. To measure the perception bias present in the continuous values, we simply computed the number of cases where ``individual A'' obtained a higher continuous value than ``individual B'', given a particular trait/dimension and subset being analysed. The analyses consist of comparing subsets of data composed of pairs of individuals with particular attributes, e.g., ``Male vs. Female'' or ``Asian vs. Caucasian'', to show how some groups were perceived differently, in some cases, as a function of their attributes. In Sec.~\ref{sec:biasanalysiscaucasian}, we analyse how facial attributes (i.e., face attractiveness and perceived age) influenced data labelling of the FI dataset. 

\subsection{Gender bias}\label{subsec:genderbias}
Table~\ref{tab:genderpreferences} shows the percentage of individuals perceived as being a more/less representative sample for a particular trait/dimension considering the gender attribute only, obtained directly from the pairwise labels (``PL''), and given the continuous values provided with the FI dataset (``CV'') for the same pairs\footnote{The subset of individuals perceived as more/less representative sample for a trait is shown in tones of red/blue, respectively (from Table~\ref{tab:genderpreferences} to Table~\ref{tab:caucvsafrican}). Differences $\geq$ 10\% are shown in bold. Note, the ``CV'' sum may not be 100\% as some pairs received the same continuous value.}. The percentages shown in Table~\ref{tab:genderpreferences} are obtained from the subset of ``valid pairs'' where individuals being compared have different gender. As it can be seen, there is an overall bias towards women, which is stronger for some traits (e.g., ``O'', ``C'' and ``E''). Interestingly, the bias is amplified for all variables during the conversion from pairwise data to continuous values using~\cite{Chen2016}. Therefore, some traits are more impacted than others.

\begin{table}[htbp]
\centering
\setlength\tabcolsep{2.8pt}
\scriptsize
\caption{Gender bias (``\textbf{M}ale vs. \textbf{F}emale'' subset), measured on the pairwise labels (PL) and continuous values (CV) provided with the FI dataset. It can be observed an overall bias towards women. Moreover, differences are amplified when data is converted from pairwise labels to continuous values.}
\begin{tabular}{lcccccccccccc}
\hline
\multicolumn{1}{|l|}{} & \multicolumn{2}{c|}{\textbf{O}} & \multicolumn{2}{c|}{\textbf{C}} & \multicolumn{2}{c|}{\textbf{E}} & \multicolumn{2}{c|}{\textbf{A}} & \multicolumn{2}{c|}{\textbf{$\overline{\mbox{N}}$}} & \multicolumn{2}{c|}{\textbf{Interview}} \\ \hline
\multicolumn{1}{|l|}{\textbf{}} & \multicolumn{1}{c|}{M} & \multicolumn{1}{c|}{F} & \multicolumn{1}{c|}{M} & \multicolumn{1}{c|}{F} & \multicolumn{1}{c|}{M} & \multicolumn{1}{c|}{F} & \multicolumn{1}{c|}{M} & \multicolumn{1}{c|}{F} & \multicolumn{1}{c|}{M} & \multicolumn{1}{c|}{F} & \multicolumn{1}{c|}{M} & \multicolumn{1}{c|}{F} \\ \hline
\multicolumn{1}{|l|}{\textbf{PL}} & \multicolumn{1}{c|}{\cellcolor[HTML]{DAE8FC}46.4} & \multicolumn{1}{c|}{\cellcolor[HTML]{FFCCC9}53.6} & \multicolumn{1}{c|}{\cellcolor[HTML]{DAE8FC}47.9} & \multicolumn{1}{c|}{\cellcolor[HTML]{FFCCC9}52.1} & \multicolumn{1}{c|}{\cellcolor[HTML]{DAE8FC}\textbf{44.7}} & \multicolumn{1}{c|}{\cellcolor[HTML]{FFCCC9}\textbf{55.3}} & \multicolumn{1}{c|}{\cellcolor[HTML]{FFCCC9}50.3} & \multicolumn{1}{c|}{\cellcolor[HTML]{DAE8FC}49.7} & \multicolumn{1}{c|}{\cellcolor[HTML]{DAE8FC}48.6} & \multicolumn{1}{c|}{\cellcolor[HTML]{FFCCC9}51.4} & \multicolumn{1}{c|}{\cellcolor[HTML]{DAE8FC}48.2} & \multicolumn{1}{c|}{\cellcolor[HTML]{FFCCC9}51.8} \\ \hline
\multicolumn{1}{|l|}{\textbf{CV}} & \multicolumn{1}{c|}{\cellcolor[HTML]{DAE8FC}\textbf{38.4}} & \multicolumn{1}{c|}{\cellcolor[HTML]{FFCCC9}\textbf{59.6}} & \multicolumn{1}{c|}{\cellcolor[HTML]{DAE8FC}\textbf{43.8}} & \multicolumn{1}{c|}{\cellcolor[HTML]{FFCCC9}\textbf{54.5}} & \multicolumn{1}{c|}{\cellcolor[HTML]{DAE8FC}\textbf{36.7}} & \multicolumn{1}{c|}{\cellcolor[HTML]{FFCCC9}\textbf{61.6}} & \multicolumn{1}{c|}{\cellcolor[HTML]{FFCCC9}49.3} & \multicolumn{1}{c|}{\cellcolor[HTML]{DAE8FC}48.4} & \multicolumn{1}{c|}{\cellcolor[HTML]{DAE8FC}45.4} & \multicolumn{1}{c|}{\cellcolor[HTML]{FFCCC9}52.7} & \multicolumn{1}{c|}{\cellcolor[HTML]{DAE8FC}44.7} & \multicolumn{1}{c|}{\cellcolor[HTML]{FFCCC9}53.6} \\ \hline
\end{tabular}
\label{tab:genderpreferences}
\end{table}

\subsection{Ethnicity and gender biases}\label{subsec:etn:gender}

Table~\ref{tab:ethnicity:gender} shows that gender had stronger influence than ethnicity when ``Asian vs. Caucasian'' subset is considered, and that there is an overall bias towards women, which is evidenced when pairs composed of individuals of different gender are used. On the contrary, Table~\ref{tab:asianvsafrican} and Table~\ref{tab:caucvsafrican} show that ethnicity had stronger influence than gender when subsets ``Asian vs. African-American'' and ``African-American vs. Caucasian'' are used. In these cases, Asians and Caucasians were more frequently perceived as being a more representative sample for a particular trait, compared to African-Americans, independently from the gender of the individuals. We can also observe a significantly lower number of pairs where both individuals are male, compared to other cases (especially in Table~\ref{tab:asianvsafrican}), which may also bias the analysis. Furthermore, as observed when analysing Table~\ref{tab:genderpreferences}, some biases were magnified when converting the binary labels to continuous values using~\cite{Chen2016}. As expected, it seems the biases are amplified during the conversion from ``PL'' to ``CV'' as a function of the bias in ``PL'', i.e., the higher the bias in pairwise labels, the higher will be the magnification when converted to continuous value. This effect showed to be stronger for smaller subsets.

\begin{table}[htbp]
\centering
\setlength\tabcolsep{2.8pt}
\scriptsize
\caption{Ethnicity and gender bias (``\textbf{Asi}an vs. \textbf{Cau}casian'' set), measured on the pairwise labels (PL) and continuous values (CV) provided with the FI dataset. In this case, gender showed a stronger influence than ethnicity (towards women).}
\begin{tabular}{lcccccccccccc}
\hline
\multicolumn{1}{|c|}{} & \multicolumn{2}{c|}{\textbf{O}} & \multicolumn{2}{c|}{\textbf{C}} & \multicolumn{2}{c|}{\textbf{E}} & \multicolumn{2}{c|}{\textbf{A}} & \multicolumn{2}{c|}{\textbf{$\overline{\mbox{N}}$}} & \multicolumn{2}{c|}{\textbf{Interview}} \\ \hline
\multicolumn{13}{|c|}{\textit{\textbf{Global}}} \\ \hline
\multicolumn{1}{|c|}{} & \multicolumn{1}{c|}{\textit{\textbf{Asi}}} & \multicolumn{1}{c|}{\textit{\textbf{Cau}}} & \multicolumn{1}{c|}{\textit{\textbf{Asi}}} & \multicolumn{1}{c|}{\textit{\textbf{Cau}}} & \multicolumn{1}{c|}{\textit{\textbf{Asi}}} & \multicolumn{1}{c|}{\textit{\textbf{Cau}}} & \multicolumn{1}{c|}{\textit{\textbf{Asi}}} & \multicolumn{1}{c|}{\textit{\textbf{Cau}}} & \multicolumn{1}{c|}{\textit{\textbf{Asi}}} & \multicolumn{1}{c|}{\textit{\textbf{Cau}}} & \multicolumn{1}{c|}{\textit{\textbf{Asi}}} & \multicolumn{1}{c|}{\textit{\textbf{Cau}}} \\ \hline
\multicolumn{1}{|l|}{\textbf{PL}} & \multicolumn{1}{c|}{\cellcolor[HTML]{FFCCC9}50.4} & \multicolumn{1}{c|}{\cellcolor[HTML]{DAE8FC}49.6} & \multicolumn{1}{c|}{\cellcolor[HTML]{FFCCC9}50.9} & \multicolumn{1}{c|}{\cellcolor[HTML]{DAE8FC}49.1} & \multicolumn{1}{c|}{\cellcolor[HTML]{FFCCC9}52.7} & \multicolumn{1}{c|}{\cellcolor[HTML]{DAE8FC}47.3} & \multicolumn{1}{c|}{\cellcolor[HTML]{DAE8FC}49.9} & \multicolumn{1}{c|}{\cellcolor[HTML]{FFCCC9}50.1} & \multicolumn{1}{c|}{50.0} & \multicolumn{1}{c|}{50.0} & \multicolumn{1}{c|}{\cellcolor[HTML]{FFCCC9}50.9} & \multicolumn{1}{c|}{\cellcolor[HTML]{DAE8FC}49.1} \\ \hline
\multicolumn{1}{|l|}{\textbf{CV}} & \multicolumn{1}{l|}{\cellcolor[HTML]{FFCCC9}49.2} & \multicolumn{1}{l|}{\cellcolor[HTML]{DAE8FC}48.3} & \multicolumn{1}{l|}{\cellcolor[HTML]{FFCCC9}50.7} & \multicolumn{1}{l|}{\cellcolor[HTML]{DAE8FC}47.4} & \multicolumn{1}{l|}{\cellcolor[HTML]{FFCCC9}\textbf{55.0}} & \multicolumn{1}{l|}{\cellcolor[HTML]{DAE8FC}\textbf{43.4}} & \multicolumn{1}{l|}{\cellcolor[HTML]{DAE8FC}47.8} & \multicolumn{1}{l|}{\cellcolor[HTML]{FFCCC9}50.0} & \multicolumn{1}{l|}{\cellcolor[HTML]{DAE8FC}47.9} & \multicolumn{1}{l|}{\cellcolor[HTML]{FFCCC9}50.1} & \multicolumn{1}{l|}{\cellcolor[HTML]{FFCCC9}49.8} & \multicolumn{1}{l|}{\cellcolor[HTML]{DAE8FC}48.3} \\ \hline
\multicolumn{13}{|c|}{\textit{\textbf{Male vs. Male}}} \\ \hline
\multicolumn{1}{|c|}{Tot.} & \multicolumn{2}{c|}{2431} & \multicolumn{2}{c|}{2495} & \multicolumn{2}{c|}{2522} & \multicolumn{2}{c|}{2504} & \multicolumn{2}{c|}{2517} & \multicolumn{2}{c|}{2549} \\ \hline
\multicolumn{1}{|l|}{\textbf{PL}} & \multicolumn{1}{c|}{\cellcolor[HTML]{FFCCC9}50.2} & \multicolumn{1}{c|}{\cellcolor[HTML]{DAE8FC}49.8} & \multicolumn{1}{c|}{\cellcolor[HTML]{FFCCC9}51.1} & \multicolumn{1}{c|}{\cellcolor[HTML]{DAE8FC}48.9} & \multicolumn{1}{c|}{\cellcolor[HTML]{FFCCC9}54.8} & \multicolumn{1}{c|}{\cellcolor[HTML]{DAE8FC}45.2} & \multicolumn{1}{c|}{\cellcolor[HTML]{FFCCC9}50.6} & \multicolumn{1}{c|}{\cellcolor[HTML]{DAE8FC}49.4} & \multicolumn{1}{c|}{\cellcolor[HTML]{FFCCC9}51.6} & \multicolumn{1}{c|}{\cellcolor[HTML]{DAE8FC}48.4} & \multicolumn{1}{c|}{\cellcolor[HTML]{FFCCC9}51.1} & \multicolumn{1}{c|}{\cellcolor[HTML]{DAE8FC}48.9} \\ \hline
\multicolumn{1}{|l|}{\textbf{CV}} & \multicolumn{1}{l|}{\cellcolor[HTML]{FFCCC9}48.7} & \multicolumn{1}{l|}{\cellcolor[HTML]{DAE8FC}48.5} & \multicolumn{1}{l|}{\cellcolor[HTML]{FFCCC9}51.7} & \multicolumn{1}{l|}{\cellcolor[HTML]{DAE8FC}46.2} & \multicolumn{1}{l|}{\cellcolor[HTML]{FFCCC9}\textbf{59.1}} & \multicolumn{1}{l|}{\cellcolor[HTML]{DAE8FC}\textbf{39.2}} & \multicolumn{1}{l|}{\cellcolor[HTML]{DAE8FC}48.6} & \multicolumn{1}{l|}{\cellcolor[HTML]{FFCCC9}49.1} & \multicolumn{1}{l|}{\cellcolor[HTML]{FFCCC9}50.1} & \multicolumn{1}{l|}{\cellcolor[HTML]{DAE8FC}47.8} & \multicolumn{1}{l|}{\cellcolor[HTML]{FFCCC9}50.8} & \multicolumn{1}{l|}{\cellcolor[HTML]{DAE8FC}47.0} \\ \hline
\multicolumn{13}{|c|}{\textit{\textbf{Female vs. Female}}} \\ \hline
\multicolumn{1}{|c|}{Tot.} & \multicolumn{2}{c|}{6459} & \multicolumn{2}{c|}{6633} & \multicolumn{2}{c|}{6756} & \multicolumn{2}{c|}{6730} & \multicolumn{2}{c|}{6771} & \multicolumn{2}{c|}{6810} \\ \hline
\multicolumn{1}{|l|}{\textbf{PL}} & \multicolumn{1}{c|}{\cellcolor[HTML]{DAE8FC}48.9} & \multicolumn{1}{c|}{\cellcolor[HTML]{FFCCC9}51.1} & \multicolumn{1}{c|}{\cellcolor[HTML]{FFCCC9}50.7} & \multicolumn{1}{c|}{\cellcolor[HTML]{DAE8FC}49.3} & \multicolumn{1}{c|}{\cellcolor[HTML]{FFCCC9}51.5} & \multicolumn{1}{c|}{\cellcolor[HTML]{DAE8FC}48.5} & \multicolumn{1}{c|}{\cellcolor[HTML]{FFCCC9}50.8} & \multicolumn{1}{c|}{\cellcolor[HTML]{DAE8FC}49.2} & \multicolumn{1}{c|}{\cellcolor[HTML]{DAE8FC}49.1} & \multicolumn{1}{c|}{\cellcolor[HTML]{FFCCC9}50.9} & \multicolumn{1}{c|}{\cellcolor[HTML]{FFCCC9}50.9} & \multicolumn{1}{c|}{\cellcolor[HTML]{DAE8FC}49.1} \\ \hline
\multicolumn{1}{|l|}{\textbf{CV}} & \multicolumn{1}{l|}{\cellcolor[HTML]{DAE8FC}46.7} & \multicolumn{1}{l|}{\cellcolor[HTML]{FFCCC9}50.9} & \multicolumn{1}{l|}{\cellcolor[HTML]{DAE8FC}48.6} & \multicolumn{1}{l|}{\cellcolor[HTML]{FFCCC9}49.5} & \multicolumn{1}{l|}{\cellcolor[HTML]{FFCCC9}50.7} & \multicolumn{1}{l|}{\cellcolor[HTML]{DAE8FC}47.6} & \multicolumn{1}{l|}{\cellcolor[HTML]{DAE8FC}48.5} & \multicolumn{1}{l|}{\cellcolor[HTML]{FFCCC9}49.5} & \multicolumn{1}{l|}{\cellcolor[HTML]{DAE8FC}46.6} & \multicolumn{1}{l|}{\cellcolor[HTML]{FFCCC9}51.4} & \multicolumn{1}{l|}{\cellcolor[HTML]{DAE8FC}48.5} & \multicolumn{1}{l|}{\cellcolor[HTML]{FFCCC9}49.6} \\ \hline
\multicolumn{13}{|c|}{\textit{\textbf{Asian Male vs. Caucasian Female}}} \\ \hline
\multicolumn{1}{|c|}{Tot.} & \multicolumn{2}{c|}{2561} & \multicolumn{2}{c|}{2619} & \multicolumn{2}{c|}{2697} & \multicolumn{2}{c|}{2656} & \multicolumn{2}{c|}{2707} & \multicolumn{2}{c|}{2724} \\ \hline
\multicolumn{1}{|l|}{\textbf{PL}} & \multicolumn{1}{c|}{\cellcolor[HTML]{DAE8FC}\textbf{44.9}} & \multicolumn{1}{c|}{\cellcolor[HTML]{FFCCC9}\textbf{55.1}} & \multicolumn{1}{c|}{\cellcolor[HTML]{DAE8FC}46.7} & \multicolumn{1}{c|}{\cellcolor[HTML]{FFCCC9}53.3} & \multicolumn{1}{c|}{\cellcolor[HTML]{DAE8FC}\textbf{44.5}} & \multicolumn{1}{c|}{\cellcolor[HTML]{FFCCC9}\textbf{55.5}} & \multicolumn{1}{c|}{\cellcolor[HTML]{DAE8FC}47.6} & \multicolumn{1}{c|}{\cellcolor[HTML]{FFCCC9}52.4} & \multicolumn{1}{c|}{\cellcolor[HTML]{DAE8FC}47.4} & \multicolumn{1}{c|}{\cellcolor[HTML]{FFCCC9}52.6} & \multicolumn{1}{c|}{\cellcolor[HTML]{DAE8FC}46.7} & \multicolumn{1}{c|}{\cellcolor[HTML]{FFCCC9}53.3} \\ \hline
\multicolumn{1}{|l|}{\textbf{CV}} & \multicolumn{1}{l|}{\cellcolor[HTML]{DAE8FC}\textbf{32.7}} & \multicolumn{1}{l|}{\cellcolor[HTML]{FFCCC9}\textbf{64.5}} & \multicolumn{1}{l|}{\cellcolor[HTML]{DAE8FC}\textbf{40.8}} & \multicolumn{1}{l|}{\cellcolor[HTML]{FFCCC9}\textbf{57.4}} & \multicolumn{1}{l|}{\cellcolor[HTML]{DAE8FC}\textbf{38.1}} & \multicolumn{1}{l|}{\cellcolor[HTML]{FFCCC9}\textbf{60.4}} & \multicolumn{1}{l|}{\cellcolor[HTML]{DAE8FC}44.4} & \multicolumn{1}{l|}{\cellcolor[HTML]{FFCCC9}53.5} & \multicolumn{1}{l|}{\cellcolor[HTML]{DAE8FC}\textbf{40.9}} & \multicolumn{1}{l|}{\cellcolor[HTML]{FFCCC9}\textbf{56.6}} & \multicolumn{1}{l|}{\cellcolor[HTML]{DAE8FC}\textbf{39.7}} & \multicolumn{1}{l|}{\cellcolor[HTML]{FFCCC9}\textbf{58.2}} \\ \hline
\multicolumn{13}{|c|}{\textit{\textbf{Asian Female vs. Caucasian Male}}} \\ \hline
\multicolumn{1}{|c|}{Tot.} & \multicolumn{2}{c|}{5916} & \multicolumn{2}{c|}{6023} & \multicolumn{2}{c|}{6162} & \multicolumn{2}{c|}{6121} & \multicolumn{2}{c|}{6160} & \multicolumn{2}{c|}{6181} \\ \hline
\multicolumn{1}{|l|}{\textbf{PL}} & \multicolumn{1}{c|}{\cellcolor[HTML]{FFCCC9}54.6} & \multicolumn{1}{c|}{\cellcolor[HTML]{DAE8FC}45.4} & \multicolumn{1}{c|}{\cellcolor[HTML]{FFCCC9}52.7} & \multicolumn{1}{c|}{\cellcolor[HTML]{DAE8FC}47.3} & \multicolumn{1}{c|}{\cellcolor[HTML]{FFCCC9}\textbf{56.6}} & \multicolumn{1}{c|}{\cellcolor[HTML]{DAE8FC}\textbf{43.4}} & \multicolumn{1}{c|}{\cellcolor[HTML]{DAE8FC}49.5} & \multicolumn{1}{c|}{\cellcolor[HTML]{FFCCC9}50.5} & \multicolumn{1}{c|}{\cellcolor[HTML]{FFCCC9}51.6} & \multicolumn{1}{c|}{\cellcolor[HTML]{DAE8FC}48.4} & \multicolumn{1}{c|}{\cellcolor[HTML]{FFCCC9}52.7} & \multicolumn{1}{c|}{\cellcolor[HTML]{DAE8FC}47.3} \\ \hline
\multicolumn{1}{|l|}{\textbf{CV}} & \multicolumn{1}{l|}{\cellcolor[HTML]{FFCCC9}\textbf{59.3}} & \multicolumn{1}{l|}{\cellcolor[HTML]{DAE8FC}\textbf{38.3}} & \multicolumn{1}{l|}{\cellcolor[HTML]{FFCCC9}\textbf{57.1}} & \multicolumn{1}{l|}{\cellcolor[HTML]{DAE8FC}\textbf{41.3}} & \multicolumn{1}{l|}{\cellcolor[HTML]{FFCCC9}\textbf{65.4}} & \multicolumn{1}{l|}{\cellcolor[HTML]{DAE8FC}\textbf{33.0}} & \multicolumn{1}{l|}{\cellcolor[HTML]{DAE8FC}48.2} & \multicolumn{1}{l|}{\cellcolor[HTML]{FFCCC9}49.5} & \multicolumn{1}{l|}{\cellcolor[HTML]{FFCCC9}51.6} & \multicolumn{1}{l|}{\cellcolor[HTML]{DAE8FC}46.6} & \multicolumn{1}{l|}{\cellcolor[HTML]{FFCCC9}\textbf{55.3}} & \multicolumn{1}{l|}{\cellcolor[HTML]{DAE8FC}\textbf{43.0}} \\ \hline
\end{tabular}
\label{tab:ethnicity:gender}
\end{table}

\begin{table}[htbp]
\centering
\setlength\tabcolsep{2.8pt}
\scriptsize
\caption{Ethnicity and gender bias (``\textbf{Asi}an vs. \textbf{Afr}ican-American'' subset), measured on the pairwise labels (PL) and continuous values (CV) provided with the FI dataset. In this case, ethnicity showed a stronger influence than gender, i.e., there is an overall bias towards Asian individuals.}
\begin{tabular}{lcccccccccccc}
\hline
\multicolumn{1}{|c|}{} & \multicolumn{2}{c|}{\textbf{O}} & \multicolumn{2}{c|}{\textbf{C}} & \multicolumn{2}{c|}{\textbf{E}} & \multicolumn{2}{c|}{\textbf{A}} & \multicolumn{2}{c|}{\textbf{$\overline{\mbox{N}}$}} & \multicolumn{2}{c|}{\textbf{Interview}} \\ \hline
\multicolumn{13}{|c|}{\textit{\textbf{Global}}} \\ \hline
\multicolumn{1}{|c|}{} & \multicolumn{1}{c|}{\textit{\textbf{Asi}}} & \multicolumn{1}{c|}{\textit{\textbf{Afr}}} & \multicolumn{1}{c|}{\textit{\textbf{Asi}}} & \multicolumn{1}{c|}{\textit{\textbf{Afr}}} & \multicolumn{1}{c|}{\textit{\textbf{Asi}}} & \multicolumn{1}{c|}{\textit{\textbf{Afr}}} & \multicolumn{1}{c|}{\textit{\textbf{Asi}}} & \multicolumn{1}{c|}{\textit{\textbf{Afr}}} & \multicolumn{1}{c|}{\textit{\textbf{Asi}}} & \multicolumn{1}{c|}{\textit{\textbf{Afr}}} & \multicolumn{1}{c|}{\textit{\textbf{Asi}}} & \multicolumn{1}{c|}{\textit{\textbf{Afr}}} \\ \hline
\multicolumn{1}{|l|}{\textbf{PL}} & \multicolumn{1}{c|}{\cellcolor[HTML]{FFCCC9}54.4} & \multicolumn{1}{c|}{\cellcolor[HTML]{DAE8FC}45.6} & \multicolumn{1}{c|}{\cellcolor[HTML]{FFCCC9}53.6} & \multicolumn{1}{c|}{\cellcolor[HTML]{DAE8FC}46.4} & \multicolumn{1}{c|}{\cellcolor[HTML]{FFCCC9}54.8} & \multicolumn{1}{c|}{\cellcolor[HTML]{DAE8FC}45.2} & \multicolumn{1}{c|}{\cellcolor[HTML]{FFCCC9}52.8} & \multicolumn{1}{c|}{\cellcolor[HTML]{DAE8FC}47.2} & \multicolumn{1}{c|}{\cellcolor[HTML]{FFCCC9}51.5} & \multicolumn{1}{c|}{\cellcolor[HTML]{DAE8FC}48.5} & \multicolumn{1}{c|}{\cellcolor[HTML]{FFCCC9}53.3} & \multicolumn{1}{c|}{\cellcolor[HTML]{DAE8FC}46.7} \\ \hline
\multicolumn{1}{|l|}{\textbf{CV}} & \multicolumn{1}{l|}{\cellcolor[HTML]{FFCCC9}\textbf{60.1}} & \multicolumn{1}{l|}{\cellcolor[HTML]{DAE8FC}\textbf{37.3}} & \multicolumn{1}{l|}{\cellcolor[HTML]{FFCCC9}\textbf{61.2}} & \multicolumn{1}{l|}{\cellcolor[HTML]{DAE8FC}\textbf{36.7}} & \multicolumn{1}{l|}{\cellcolor[HTML]{FFCCC9}\textbf{63.6}} & \multicolumn{1}{l|}{\cellcolor[HTML]{DAE8FC}\textbf{34.9}} & \multicolumn{1}{l|}{\cellcolor[HTML]{FFCCC9}\textbf{57.2}} & \multicolumn{1}{l|}{\cellcolor[HTML]{DAE8FC}\textbf{40.3}} & \multicolumn{1}{l|}{\cellcolor[HTML]{FFCCC9}\textbf{56.3}} & \multicolumn{1}{l|}{\cellcolor[HTML]{DAE8FC}\textbf{41.7}} & \multicolumn{1}{l|}{\cellcolor[HTML]{FFCCC9}\textbf{60.9}} & \multicolumn{1}{l|}{\cellcolor[HTML]{DAE8FC}\textbf{37.2}} \\ \hline
\multicolumn{13}{|c|}{\textit{\textbf{Male vs. Male}}} \\ \hline
\multicolumn{1}{|c|}{Tot.} & \multicolumn{2}{c|}{187} & \multicolumn{2}{c|}{191} & \multicolumn{2}{c|}{196} & \multicolumn{2}{c|}{194} & \multicolumn{2}{c|}{197} & \multicolumn{2}{c|}{204} \\ \hline
\multicolumn{1}{|l|}{\textbf{PL}} & \multicolumn{1}{c|}{\cellcolor[HTML]{FFCCC9}51.3} & \multicolumn{1}{c|}{\cellcolor[HTML]{DAE8FC}48.7} & \multicolumn{1}{c|}{\cellcolor[HTML]{FFCCC9}\textbf{55.5}} & \multicolumn{1}{c|}{\cellcolor[HTML]{DAE8FC}\textbf{44.5}} & \multicolumn{1}{c|}{\cellcolor[HTML]{FFCCC9}54.1} & \multicolumn{1}{c|}{\cellcolor[HTML]{DAE8FC}45.9} & \multicolumn{1}{c|}{\cellcolor[HTML]{FFCCC9}51.5} & \multicolumn{1}{c|}{\cellcolor[HTML]{DAE8FC}48.5} & \multicolumn{1}{c|}{\cellcolor[HTML]{DAE8FC}49.7} & \multicolumn{1}{c|}{\cellcolor[HTML]{FFCCC9}50.3} & \multicolumn{1}{c|}{\cellcolor[HTML]{FFCCC9}52.5} & \multicolumn{1}{c|}{\cellcolor[HTML]{DAE8FC}47.5} \\ \hline
\multicolumn{1}{|l|}{\textbf{CV}} & \multicolumn{1}{l|}{\cellcolor[HTML]{FFCCC9}51.9} & \multicolumn{1}{l|}{\cellcolor[HTML]{DAE8FC}46.0} & \multicolumn{1}{l|}{\cellcolor[HTML]{FFCCC9}\textbf{60.2}} & \multicolumn{1}{l|}{\cellcolor[HTML]{DAE8FC}\textbf{39.3}} & \multicolumn{1}{l|}{\cellcolor[HTML]{FFCCC9}\textbf{61.2}} & \multicolumn{1}{l|}{\cellcolor[HTML]{DAE8FC}\textbf{37.2}} & \multicolumn{1}{l|}{\cellcolor[HTML]{FFCCC9}52.6} & \multicolumn{1}{l|}{\cellcolor[HTML]{DAE8FC}44.3} & \multicolumn{1}{l|}{\cellcolor[HTML]{FFCCC9}48.7} & \multicolumn{1}{l|}{\cellcolor[HTML]{DAE8FC}46.2} & \multicolumn{1}{l|}{\cellcolor[HTML]{FFCCC9}\textbf{54.4}} & \multicolumn{1}{l|}{\cellcolor[HTML]{DAE8FC}\textbf{43.6}} \\ \hline
\multicolumn{13}{|c|}{\textit{\textbf{Female vs. Female}}} \\ \hline
\multicolumn{1}{|c|}{Tot.} & \multicolumn{2}{c|}{1051} & \multicolumn{2}{c|}{1077} & \multicolumn{2}{c|}{1124} & \multicolumn{2}{c|}{1096} & \multicolumn{2}{c|}{1113} & \multicolumn{2}{c|}{1118} \\ \hline
\multicolumn{1}{|l|}{\textbf{PL}} & \multicolumn{1}{c|}{\cellcolor[HTML]{FFCCC9}54.5} & \multicolumn{1}{c|}{\cellcolor[HTML]{DAE8FC}45.5} & \multicolumn{1}{c|}{\cellcolor[HTML]{FFCCC9}54.2} & \multicolumn{1}{c|}{\cellcolor[HTML]{DAE8FC}45.8} & \multicolumn{1}{c|}{\cellcolor[HTML]{FFCCC9}54.0} & \multicolumn{1}{c|}{\cellcolor[HTML]{DAE8FC}46.0} & \multicolumn{1}{c|}{\cellcolor[HTML]{FFCCC9}52.6} & \multicolumn{1}{c|}{\cellcolor[HTML]{DAE8FC}47.4} & \multicolumn{1}{c|}{\cellcolor[HTML]{FFCCC9}52.3} & \multicolumn{1}{c|}{\cellcolor[HTML]{DAE8FC}47.7} & \multicolumn{1}{c|}{\cellcolor[HTML]{FFCCC9}53.8} & \multicolumn{1}{c|}{\cellcolor[HTML]{DAE8FC}46.2} \\ \hline
\multicolumn{1}{|l|}{\textbf{CV}} & \multicolumn{1}{l|}{\cellcolor[HTML]{FFCCC9}\textbf{64.5}} & \multicolumn{1}{l|}{\cellcolor[HTML]{DAE8FC}\textbf{32.7}} & \multicolumn{1}{l|}{\cellcolor[HTML]{FFCCC9}\textbf{63.1}} & \multicolumn{1}{l|}{\cellcolor[HTML]{DAE8FC}\textbf{34.4}} & \multicolumn{1}{l|}{\cellcolor[HTML]{FFCCC9}\textbf{66.6}} & \multicolumn{1}{l|}{\cellcolor[HTML]{DAE8FC}\textbf{32.0}} & \multicolumn{1}{l|}{\cellcolor[HTML]{FFCCC9}\textbf{61.0}} & \multicolumn{1}{l|}{\cellcolor[HTML]{DAE8FC}\textbf{36.1}} & \multicolumn{1}{l|}{\cellcolor[HTML]{FFCCC9}\textbf{61.0}} & \multicolumn{1}{l|}{\cellcolor[HTML]{DAE8FC}\textbf{37.8}} & \multicolumn{1}{l|}{\cellcolor[HTML]{FFCCC9}\textbf{64.7}} & \multicolumn{1}{l|}{\cellcolor[HTML]{DAE8FC}\textbf{33.3}} \\ \hline
\multicolumn{13}{|c|}{\textit{\textbf{Asian Male vs. African-American Female}}} \\ \hline
\multicolumn{1}{|c|}{Tot.} & \multicolumn{2}{c|}{428} & \multicolumn{2}{c|}{439} & \multicolumn{2}{c|}{460} & \multicolumn{2}{c|}{457} & \multicolumn{2}{c|}{451} & \multicolumn{2}{c|}{470} \\ \hline
\multicolumn{1}{|l|}{\textbf{PL}} & \multicolumn{1}{c|}{\cellcolor[HTML]{FFCCC9}\textbf{56.5}} & \multicolumn{1}{c|}{\cellcolor[HTML]{DAE8FC}\textbf{43.5}} & \multicolumn{1}{c|}{\cellcolor[HTML]{FFCCC9}\textbf{56.5}} & \multicolumn{1}{c|}{\cellcolor[HTML]{DAE8FC}\textbf{43.5}} & \multicolumn{1}{c|}{\cellcolor[HTML]{FFCCC9}\textbf{57.2}} & \multicolumn{1}{c|}{\cellcolor[HTML]{DAE8FC}\textbf{42.8}} & \multicolumn{1}{c|}{\cellcolor[HTML]{FFCCC9}\textbf{57.3}} & \multicolumn{1}{c|}{\cellcolor[HTML]{DAE8FC}\textbf{42.7}} & \multicolumn{1}{c|}{\cellcolor[HTML]{FFCCC9}52.8} & \multicolumn{1}{c|}{\cellcolor[HTML]{DAE8FC}47.2} & \multicolumn{1}{c|}{\cellcolor[HTML]{FFCCC9}\textbf{56.0}} & \multicolumn{1}{c|}{\cellcolor[HTML]{DAE8FC}\textbf{44.0}} \\ \hline
\multicolumn{1}{|l|}{\textbf{CV}} & \multicolumn{1}{l|}{\cellcolor[HTML]{DAE8FC}48.6} & \multicolumn{1}{l|}{\cellcolor[HTML]{FFCCC9}48.8} & \multicolumn{1}{l|}{\cellcolor[HTML]{FFCCC9}\textbf{58.3}} & \multicolumn{1}{l|}{\cellcolor[HTML]{DAE8FC}\textbf{39.9}} & \multicolumn{1}{l|}{\cellcolor[HTML]{FFCCC9}\textbf{54.6}} & \multicolumn{1}{l|}{\cellcolor[HTML]{DAE8FC}\textbf{43.7}} & \multicolumn{1}{l|}{\cellcolor[HTML]{FFCCC9}\textbf{58.4}} & \multicolumn{1}{l|}{\cellcolor[HTML]{DAE8FC}\textbf{39.4}} & \multicolumn{1}{l|}{\cellcolor[HTML]{FFCCC9}\textbf{55.2}} & \multicolumn{1}{l|}{\cellcolor[HTML]{DAE8FC}\textbf{42.1}} & \multicolumn{1}{l|}{\cellcolor[HTML]{FFCCC9}\textbf{56.6}} & \multicolumn{1}{l|}{\cellcolor[HTML]{DAE8FC}\textbf{41.1}} \\ \hline
\multicolumn{13}{|c|}{\textit{\textbf{Asian Female vs. African-American Male}}} \\ \hline
\multicolumn{1}{|c|}{Tot.} & \multicolumn{2}{c|}{538} & \multicolumn{2}{c|}{554} & \multicolumn{2}{c|}{550} & \multicolumn{2}{c|}{549} & \multicolumn{2}{c|}{550} & \multicolumn{2}{c|}{549} \\ \hline
\multicolumn{1}{|l|}{\textbf{PL}} & \multicolumn{1}{c|}{\cellcolor[HTML]{FFCCC9}53.4} & \multicolumn{1}{c|}{\cellcolor[HTML]{DAE8FC}46.6} & \multicolumn{1}{c|}{\cellcolor[HTML]{DAE8FC}49.5} & \multicolumn{1}{c|}{\cellcolor[HTML]{FFCCC9}50.5} & \multicolumn{1}{c|}{\cellcolor[HTML]{FFCCC9}54.7} & \multicolumn{1}{c|}{\cellcolor[HTML]{DAE8FC}45.3} & \multicolumn{1}{c|}{\cellcolor[HTML]{DAE8FC}49.9} & \multicolumn{1}{c|}{\cellcolor[HTML]{FFCCC9}50.1} & \multicolumn{1}{c|}{\cellcolor[HTML]{DAE8FC}49.4} & \multicolumn{1}{c|}{\cellcolor[HTML]{FFCCC9}50.6} & \multicolumn{1}{c|}{\cellcolor[HTML]{FFCCC9}50.3} & \multicolumn{1}{c|}{\cellcolor[HTML]{DAE8FC}49.7} \\ \hline
\multicolumn{1}{|l|}{\textbf{CV}} & \multicolumn{1}{l|}{\cellcolor[HTML]{FFCCC9}\textbf{63.6}} & \multicolumn{1}{l|}{\cellcolor[HTML]{DAE8FC}\textbf{34.0}} & \multicolumn{1}{l|}{\cellcolor[HTML]{FFCCC9}\textbf{60.1}} & \multicolumn{1}{l|}{\cellcolor[HTML]{DAE8FC}\textbf{37.9}} & \multicolumn{1}{l|}{\cellcolor[HTML]{FFCCC9}\textbf{65.6}} & \multicolumn{1}{l|}{\cellcolor[HTML]{DAE8FC}\textbf{32.7}} & \multicolumn{1}{l|}{\cellcolor[HTML]{FFCCC9}50.1} & \multicolumn{1}{l|}{\cellcolor[HTML]{DAE8FC}47.9} & \multicolumn{1}{l|}{\cellcolor[HTML]{FFCCC9}50.4} & \multicolumn{1}{l|}{\cellcolor[HTML]{DAE8FC}47.6} & \multicolumn{1}{l|}{\cellcolor[HTML]{FFCCC9}\textbf{59.2}} & \multicolumn{1}{l|}{\cellcolor[HTML]{DAE8FC}\textbf{39.5}} \\ \hline
\end{tabular}
\label{tab:asianvsafrican}
\end{table}

\begin{table}[htbp]
\centering
\setlength\tabcolsep{2.8pt}
\scriptsize
\caption{Ethnicity and gender bias (``\textbf{Afr}ican-American vs. \textbf{Cau}casian'' subset), measured on the pairwise labels (PL) and continuous values (CV) provided with the FI dataset. In this case, ethnicity showed a stronger influence than gender, i.e., there is an overall bias towards Caucasian individuals.}
\begin{tabular}{lcccccccccccc}
\hline
\multicolumn{1}{|c|}{} & \multicolumn{2}{c|}{\textbf{O}} & \multicolumn{2}{c|}{\textbf{C}} & \multicolumn{2}{c|}{\textbf{E}} & \multicolumn{2}{c|}{\textbf{A}} & \multicolumn{2}{c|}{\textbf{$\overline{\mbox{N}}$}} & \multicolumn{2}{c|}{\textbf{Interview}} \\ \hline
\multicolumn{13}{|c|}{\textit{\textbf{Global}}} \\ \hline
\multicolumn{1}{|c|}{} & \multicolumn{1}{c|}{\textit{\textbf{Afr}}} & \multicolumn{1}{c|}{\textit{\textbf{Cau}}} & \multicolumn{1}{c|}{\textit{\textbf{Afr}}} & \multicolumn{1}{c|}{\textit{\textbf{Cau}}} & \multicolumn{1}{c|}{\textit{\textbf{Afr}}} & \multicolumn{1}{c|}{\textit{\textbf{Cau}}} & \multicolumn{1}{c|}{\textit{\textbf{Afr}}} & \multicolumn{1}{c|}{\textit{\textbf{Cau}}} & \multicolumn{1}{c|}{\textit{\textbf{Afr}}} & \multicolumn{1}{c|}{\textit{\textbf{Cau}}} & \multicolumn{1}{c|}{\textit{\textbf{Afr}}} & \multicolumn{1}{c|}{\textit{\textbf{Cau}}} \\ \hline
\multicolumn{1}{|l|}{\textbf{PL}} & \multicolumn{1}{c|}{\cellcolor[HTML]{DAE8FC}46.4} & \multicolumn{1}{c|}{\cellcolor[HTML]{FFCCC9}53.6} & \multicolumn{1}{c|}{\cellcolor[HTML]{DAE8FC}46.8} & \multicolumn{1}{c|}{\cellcolor[HTML]{FFCCC9}53.2} & \multicolumn{1}{c|}{\cellcolor[HTML]{DAE8FC}47.1} & \multicolumn{1}{c|}{\cellcolor[HTML]{FFCCC9}52.9} & \multicolumn{1}{c|}{\cellcolor[HTML]{DAE8FC}47.6} & \multicolumn{1}{c|}{\cellcolor[HTML]{FFCCC9}52.4} & \multicolumn{1}{c|}{\cellcolor[HTML]{DAE8FC}47.8} & \multicolumn{1}{c|}{\cellcolor[HTML]{FFCCC9}52.2} & \multicolumn{1}{c|}{\cellcolor[HTML]{DAE8FC}47.0} & \multicolumn{1}{c|}{\cellcolor[HTML]{FFCCC9}53.0} \\ \hline
\multicolumn{1}{|l|}{\textbf{CV}} & \multicolumn{1}{l|}{\cellcolor[HTML]{DAE8FC}\textbf{38.7}} & \multicolumn{1}{l|}{\cellcolor[HTML]{FFCCC9}\textbf{59.1}} & \multicolumn{1}{l|}{\cellcolor[HTML]{DAE8FC}\textbf{41.7}} & \multicolumn{1}{l|}{\cellcolor[HTML]{FFCCC9}\textbf{56.6}} & \multicolumn{1}{l|}{\cellcolor[HTML]{DAE8FC}\textbf{42.5}} & \multicolumn{1}{l|}{\cellcolor[HTML]{FFCCC9}\textbf{55.9}} & \multicolumn{1}{l|}{\cellcolor[HTML]{DAE8FC}\textbf{41.0}} & \multicolumn{1}{l|}{\cellcolor[HTML]{FFCCC9}\textbf{56.6}} & \multicolumn{1}{l|}{\cellcolor[HTML]{DAE8FC}\textbf{42.7}} & \multicolumn{1}{l|}{\cellcolor[HTML]{FFCCC9}\textbf{55.5}} & \multicolumn{1}{l|}{\cellcolor[HTML]{DAE8FC}\textbf{41.4}} & \multicolumn{1}{l|}{\cellcolor[HTML]{FFCCC9}\textbf{56.8}} \\ \hline
\multicolumn{13}{|c|}{\textit{\textbf{Male vs. Male}}} \\ \hline
\multicolumn{1}{|c|}{Tot.} & \multicolumn{2}{c|}{8364} & \multicolumn{2}{c|}{8559} & \multicolumn{2}{c|}{8726} & \multicolumn{2}{c|}{8655} & \multicolumn{2}{c|}{8737} & \multicolumn{2}{c|}{8792} \\ \hline
\multicolumn{1}{|l|}{\textbf{PL}} & \multicolumn{1}{c|}{\cellcolor[HTML]{DAE8FC}47.7} & \multicolumn{1}{c|}{\cellcolor[HTML]{FFCCC9}52.3} & \multicolumn{1}{c|}{\cellcolor[HTML]{DAE8FC}48.7} & \multicolumn{1}{c|}{\cellcolor[HTML]{FFCCC9}51.3} & \multicolumn{1}{c|}{\cellcolor[HTML]{DAE8FC}48.7} & \multicolumn{1}{c|}{\cellcolor[HTML]{FFCCC9}51.3} & \multicolumn{1}{c|}{\cellcolor[HTML]{DAE8FC}49.1} & \multicolumn{1}{c|}{\cellcolor[HTML]{FFCCC9}50.9} & \multicolumn{1}{c|}{\cellcolor[HTML]{DAE8FC}49.6} & \multicolumn{1}{c|}{\cellcolor[HTML]{FFCCC9}50.4} & \multicolumn{1}{c|}{\cellcolor[HTML]{DAE8FC}48.8} & \multicolumn{1}{c|}{\cellcolor[HTML]{FFCCC9}51.2} \\ \hline
\multicolumn{1}{|l|}{\textbf{CV}} & \multicolumn{1}{l|}{\cellcolor[HTML]{DAE8FC}\textbf{43.5}} & \multicolumn{1}{l|}{\cellcolor[HTML]{FFCCC9}\textbf{54.3}} & \multicolumn{1}{l|}{\cellcolor[HTML]{DAE8FC}46.5} & \multicolumn{1}{l|}{\cellcolor[HTML]{FFCCC9}51.9} & \multicolumn{1}{l|}{\cellcolor[HTML]{DAE8FC}45.9} & \multicolumn{1}{l|}{\cellcolor[HTML]{FFCCC9}52.3} & \multicolumn{1}{l|}{\cellcolor[HTML]{DAE8FC}47.1} & \multicolumn{1}{l|}{\cellcolor[HTML]{FFCCC9}50.5} & \multicolumn{1}{l|}{49.1} & \multicolumn{1}{l|}{49.1} & \multicolumn{1}{l|}{\cellcolor[HTML]{DAE8FC}47.0} & \multicolumn{1}{l|}{\cellcolor[HTML]{FFCCC9}51.3} \\ \hline
\multicolumn{13}{|c|}{\textit{\textbf{Female vs. Female}}} \\ \hline
\multicolumn{1}{|c|}{Tot.} & \multicolumn{2}{c|}{20339} & \multicolumn{2}{c|}{20667} & \multicolumn{2}{c|}{21254} & \multicolumn{2}{c|}{21050} & \multicolumn{2}{c|}{21221} & \multicolumn{2}{c|}{21347} \\ \hline
\multicolumn{1}{|l|}{\textbf{PL}} & \multicolumn{1}{c|}{\cellcolor[HTML]{DAE8FC}\textbf{44.7}} & \multicolumn{1}{c|}{\cellcolor[HTML]{FFCCC9}\textbf{55.3}} & \multicolumn{1}{c|}{\cellcolor[HTML]{DAE8FC}\textbf{45.4}} & \multicolumn{1}{c|}{\cellcolor[HTML]{FFCCC9}\textbf{54.6}} & \multicolumn{1}{c|}{\cellcolor[HTML]{DAE8FC}\textbf{44.9}} & \multicolumn{1}{c|}{\cellcolor[HTML]{FFCCC9}\textbf{55.1}} & \multicolumn{1}{c|}{\cellcolor[HTML]{DAE8FC}46.9} & \multicolumn{1}{c|}{\cellcolor[HTML]{FFCCC9}53.1} & \multicolumn{1}{c|}{\cellcolor[HTML]{DAE8FC}46.6} & \multicolumn{1}{c|}{\cellcolor[HTML]{FFCCC9}53.4} & \multicolumn{1}{c|}{\cellcolor[HTML]{DAE8FC}45.6} & \multicolumn{1}{c|}{\cellcolor[HTML]{FFCCC9}54.4} \\ \hline
\multicolumn{1}{|l|}{\textbf{CV}} & \multicolumn{1}{l|}{\cellcolor[HTML]{DAE8FC}\textbf{33.8}} & \multicolumn{1}{l|}{\cellcolor[HTML]{FFCCC9}\textbf{64.3}} & \multicolumn{1}{l|}{\cellcolor[HTML]{DAE8FC}\textbf{37.7}} & \multicolumn{1}{l|}{\cellcolor[HTML]{FFCCC9}\textbf{60.5}} & \multicolumn{1}{l|}{\cellcolor[HTML]{DAE8FC}\textbf{37.4}} & \multicolumn{1}{l|}{\cellcolor[HTML]{FFCCC9}\textbf{61.0}} & \multicolumn{1}{l|}{\cellcolor[HTML]{DAE8FC}\textbf{38.4}} & \multicolumn{1}{l|}{\cellcolor[HTML]{FFCCC9}\textbf{59.3}} & \multicolumn{1}{l|}{\cellcolor[HTML]{DAE8FC}\textbf{38.7}} & \multicolumn{1}{l|}{\cellcolor[HTML]{FFCCC9}\textbf{59.6}} & \multicolumn{1}{l|}{\cellcolor[HTML]{DAE8FC}\textbf{37.7}} & \multicolumn{1}{l|}{\cellcolor[HTML]{FFCCC9}\textbf{60.5}} \\ \hline
\multicolumn{13}{|c|}{\textit{\textbf{African-American Male vs. Caucasian Female}}} \\ \hline
\multicolumn{1}{|c|}{Tot.} & \multicolumn{2}{c|}{9411} & \multicolumn{2}{c|}{9594} & \multicolumn{2}{c|}{9865} & \multicolumn{2}{c|}{9792} & \multicolumn{2}{c|}{9853} & \multicolumn{2}{c|}{9915} \\ \hline
\multicolumn{1}{|l|}{\textbf{PL}} & \multicolumn{1}{c|}{\cellcolor[HTML]{DAE8FC}\textbf{44.3}} & \multicolumn{1}{c|}{\cellcolor[HTML]{FFCCC9}\textbf{55.7}} & \multicolumn{1}{c|}{\cellcolor[HTML]{DAE8FC}46.5} & \multicolumn{1}{c|}{\cellcolor[HTML]{FFCCC9}53.5} & \multicolumn{1}{c|}{\cellcolor[HTML]{DAE8FC}\textbf{43.1}} & \multicolumn{1}{c|}{\cellcolor[HTML]{FFCCC9}\textbf{56.9}} & \multicolumn{1}{c|}{\cellcolor[HTML]{DAE8FC}49.9} & \multicolumn{1}{c|}{\cellcolor[HTML]{FFCCC9}50.1} & \multicolumn{1}{c|}{\cellcolor[HTML]{DAE8FC}48.7} & \multicolumn{1}{c|}{\cellcolor[HTML]{FFCCC9}51.3} & \multicolumn{1}{c|}{\cellcolor[HTML]{DAE8FC}47.4} & \multicolumn{1}{c|}{\cellcolor[HTML]{FFCCC9}52.6} \\ \hline
\multicolumn{1}{|l|}{\textbf{CV}} & \multicolumn{1}{l|}{\cellcolor[HTML]{DAE8FC}\textbf{32.3}} & \multicolumn{1}{l|}{\cellcolor[HTML]{FFCCC9}\textbf{65.6}} & \multicolumn{1}{l|}{\cellcolor[HTML]{DAE8FC}\textbf{40.4}} & \multicolumn{1}{l|}{\cellcolor[HTML]{FFCCC9}\textbf{57.8}} & \multicolumn{1}{l|}{\cellcolor[HTML]{DAE8FC}\textbf{33.8}} & \multicolumn{1}{l|}{\cellcolor[HTML]{FFCCC9}\textbf{64.8}} & \multicolumn{1}{l|}{\cellcolor[HTML]{DAE8FC}46.2} & \multicolumn{1}{l|}{\cellcolor[HTML]{FFCCC9}51.5} & \multicolumn{1}{l|}{\cellcolor[HTML]{DAE8FC}44.3} & \multicolumn{1}{l|}{\cellcolor[HTML]{FFCCC9}53.8} & \multicolumn{1}{l|}{\cellcolor[HTML]{DAE8FC}\textbf{41.6}} & \multicolumn{1}{l|}{\cellcolor[HTML]{FFCCC9}\textbf{56.8}} \\ \hline
\multicolumn{13}{|c|}{\textit{\textbf{African-American Female vs. Caucasian Male}}} \\ \hline
\multicolumn{1}{|c|}{Tot.} & \multicolumn{2}{c|}{18382} & \multicolumn{2}{c|}{18766} & \multicolumn{2}{c|}{19255} & \multicolumn{2}{c|}{19046} & \multicolumn{2}{c|}{19197} & \multicolumn{2}{c|}{19333} \\ \hline
\multicolumn{1}{|l|}{\textbf{PL}} & \multicolumn{1}{c|}{\cellcolor[HTML]{DAE8FC}48.6} & \multicolumn{1}{c|}{\cellcolor[HTML]{FFCCC9}51.4} & \multicolumn{1}{c|}{\cellcolor[HTML]{DAE8FC}47.7} & \multicolumn{1}{c|}{\cellcolor[HTML]{FFCCC9}52.3} & \multicolumn{1}{c|}{\cellcolor[HTML]{FFCCC9}50.9} & \multicolumn{1}{c|}{\cellcolor[HTML]{DAE8FC}49.1} & \multicolumn{1}{c|}{\cellcolor[HTML]{DAE8FC}46.6} & \multicolumn{1}{c|}{\cellcolor[HTML]{FFCCC9}53.4} & \multicolumn{1}{c|}{\cellcolor[HTML]{DAE8FC}47.7} & \multicolumn{1}{c|}{\cellcolor[HTML]{FFCCC9}52.3} & \multicolumn{1}{c|}{\cellcolor[HTML]{DAE8FC}47.5} & \multicolumn{1}{c|}{\cellcolor[HTML]{FFCCC9}52.5} \\ \hline
\multicolumn{1}{|l|}{\textbf{CV}} & \multicolumn{1}{l|}{\cellcolor[HTML]{DAE8FC}45.4} & \multicolumn{1}{l|}{\cellcolor[HTML]{FFCCC9}52.3} & \multicolumn{1}{l|}{\cellcolor[HTML]{DAE8FC}44.5} & \multicolumn{1}{l|}{\cellcolor[HTML]{FFCCC9}53.7} & \multicolumn{1}{l|}{\cellcolor[HTML]{FFCCC9}50.9} & \multicolumn{1}{l|}{\cellcolor[HTML]{DAE8FC}47.3} & \multicolumn{1}{l|}{\cellcolor[HTML]{DAE8FC}\textbf{38.5}} & \multicolumn{1}{l|}{\cellcolor[HTML]{FFCCC9}\textbf{59.1}} & \multicolumn{1}{l|}{\cellcolor[HTML]{DAE8FC}\textbf{43.2}} & \multicolumn{1}{l|}{\cellcolor[HTML]{FFCCC9}\textbf{54.8}} & \multicolumn{1}{l|}{\cellcolor[HTML]{DAE8FC}\textbf{42.9}} & \multicolumn{1}{l|}{\cellcolor[HTML]{FFCCC9}\textbf{55.2}} \\ \hline
\end{tabular}
\label{tab:caucvsafrican}
\end{table}

\subsection{Face attributes and related biases}\label{sec:biasanalysiscaucasian}
\begin{figure}[htbp]
	\centering
	\begin{subfigure}
	{\includegraphics[width=0.48\linewidth]{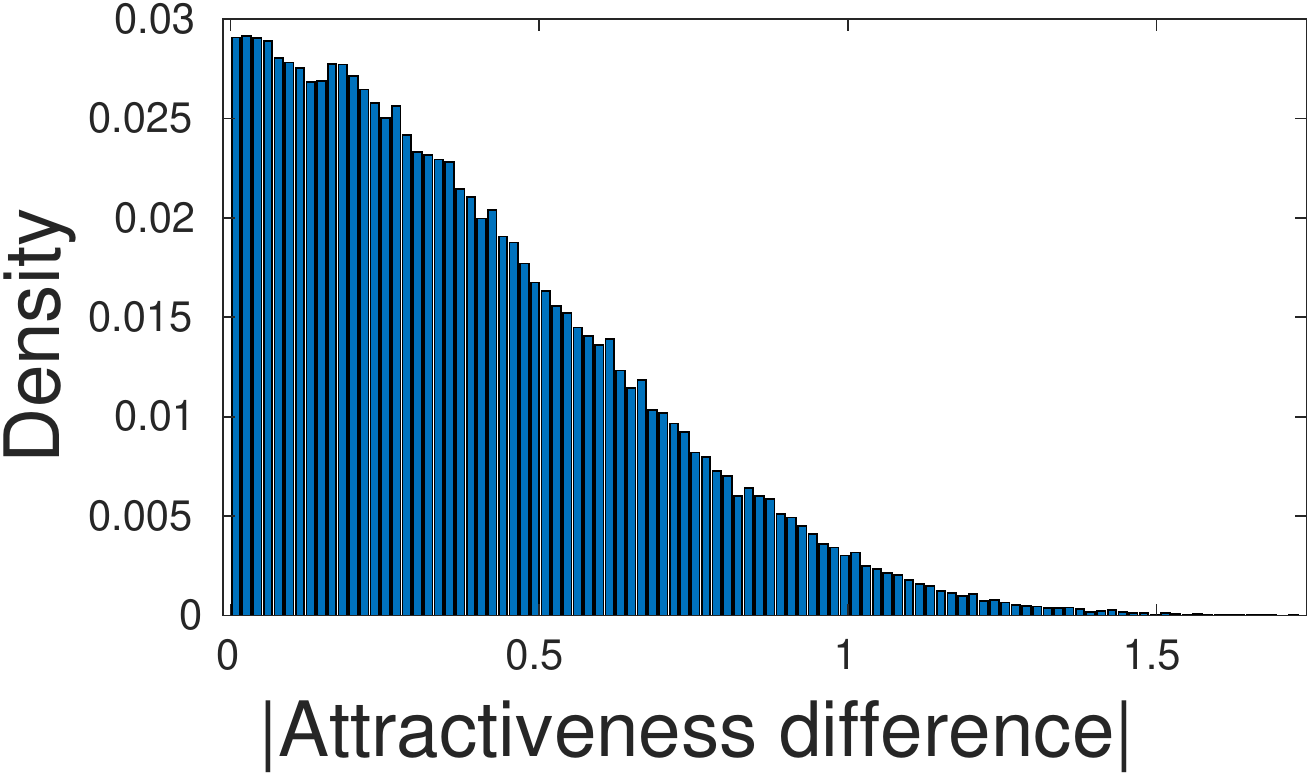}}
	\end{subfigure}
	\begin{subfigure}
	{\includegraphics[width=0.48\linewidth]{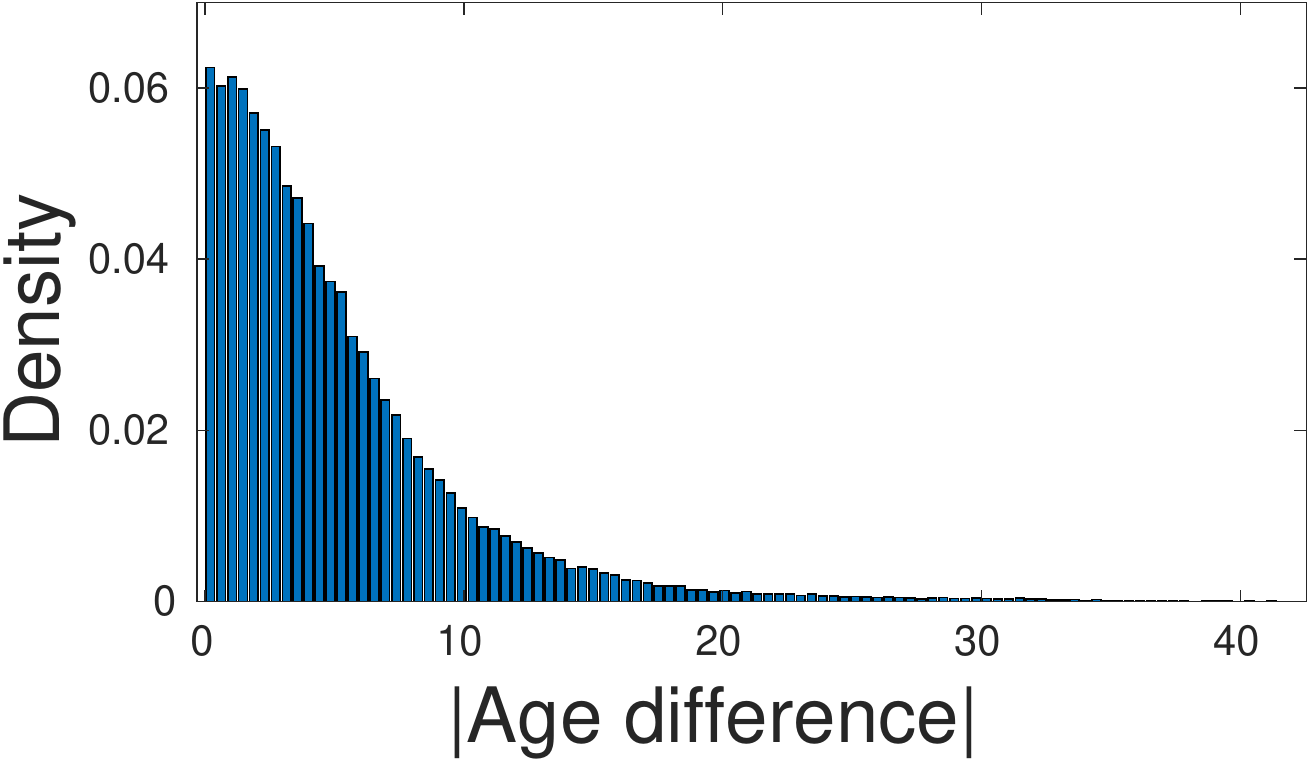}}
	\end{subfigure}
\caption{Distributions of face attribute differences between individuals in a pair (``Caucasian vs. Caucasian'' subset, pairs composed of individuals of same gender).}	
	\label{fig:att_age_diff}
\end{figure}

Given the face attributes extracted for all Caucasian individuals (detailed in Sec.~\ref{sec:faceattributeextraction}), we are able to analyse their influence on data labelling of the FI dataset. To remove the gender variable from the analysis, only pairs of individuals having the same gender are considered. For the sake of illustration, Fig.~\ref{fig:att_age_diff} shows the distributions of face attribute differences between pairs of individuals in this subset.  

\begin{figure*}[htbp]
	\centering
	\begin{subfigure}
	{\includegraphics[width=0.47\linewidth]{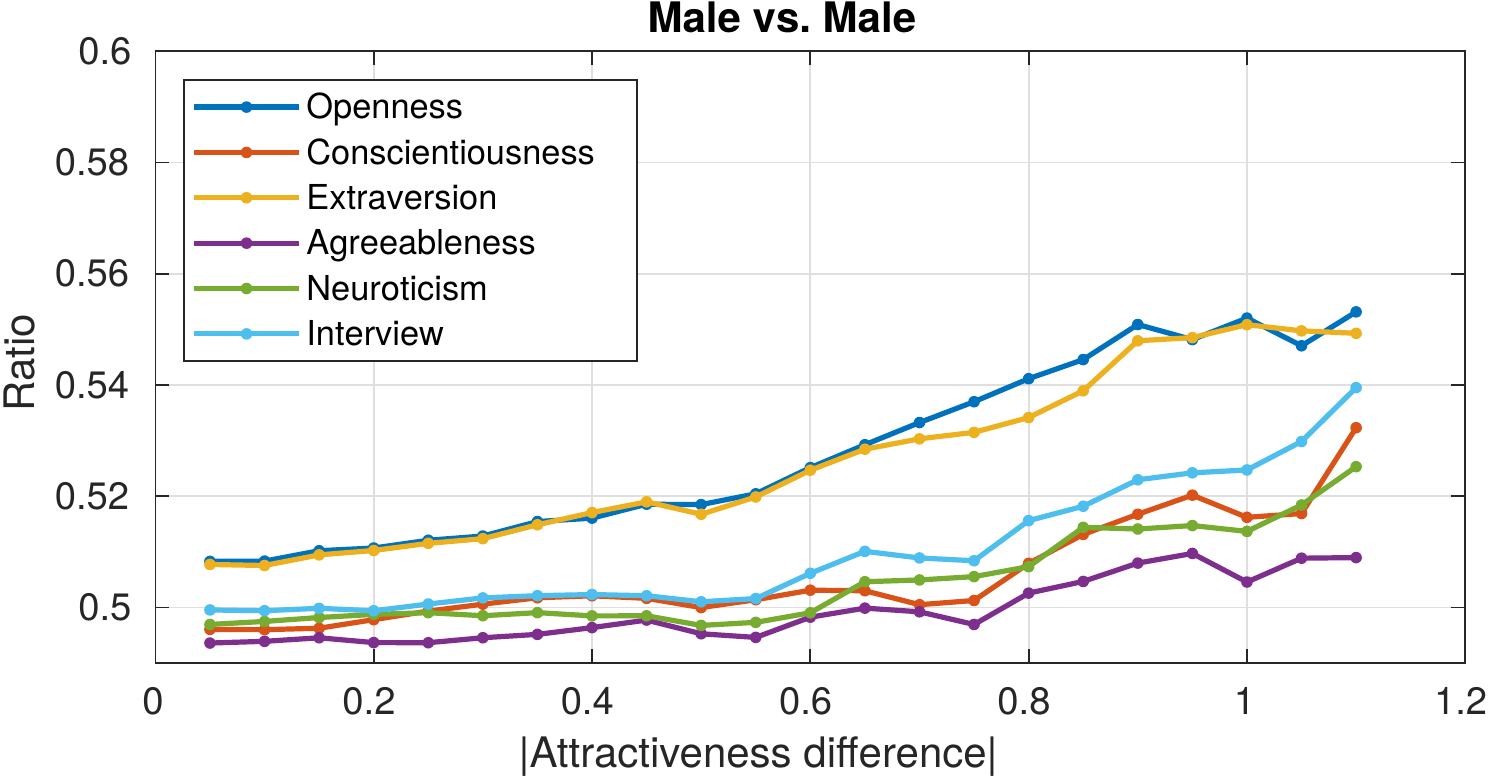}}
	\end{subfigure}
	\begin{subfigure}
	{\includegraphics[width=0.47\linewidth]{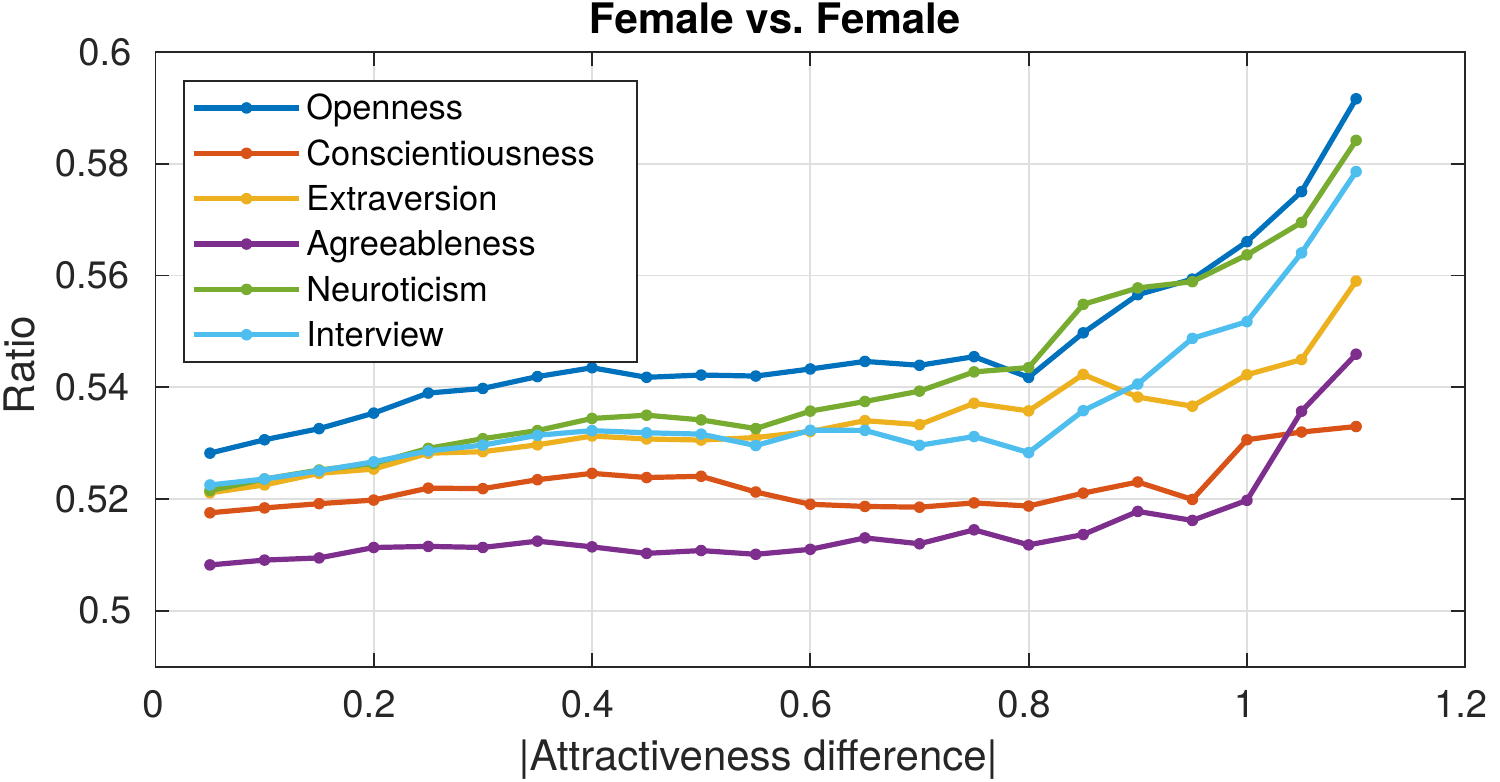}}
	\end{subfigure}
\caption{Face attractiveness bias (``Caucasian vs. Caucasian'' subset). Number of times an individual in a pair recognised as ``more attractive'' was chosen, divided by the number of times a ``less attractive'' individual was selected, as a function of the attractiveness difference between them. Ratio $> 0.5$ indicates that ``more attractive'' individuals are more frequently selected, a trend that can be clearly seen from these plots, especially for larger differences. Note that in this case, \textit{Neuroticism} relates to ``Emotion stability''.}	
	\label{fig:att_th_bias}
\end{figure*}

\begin{figure*}[htbp]
	\centering
	\begin{subfigure}
	{\includegraphics[width=0.46\linewidth]{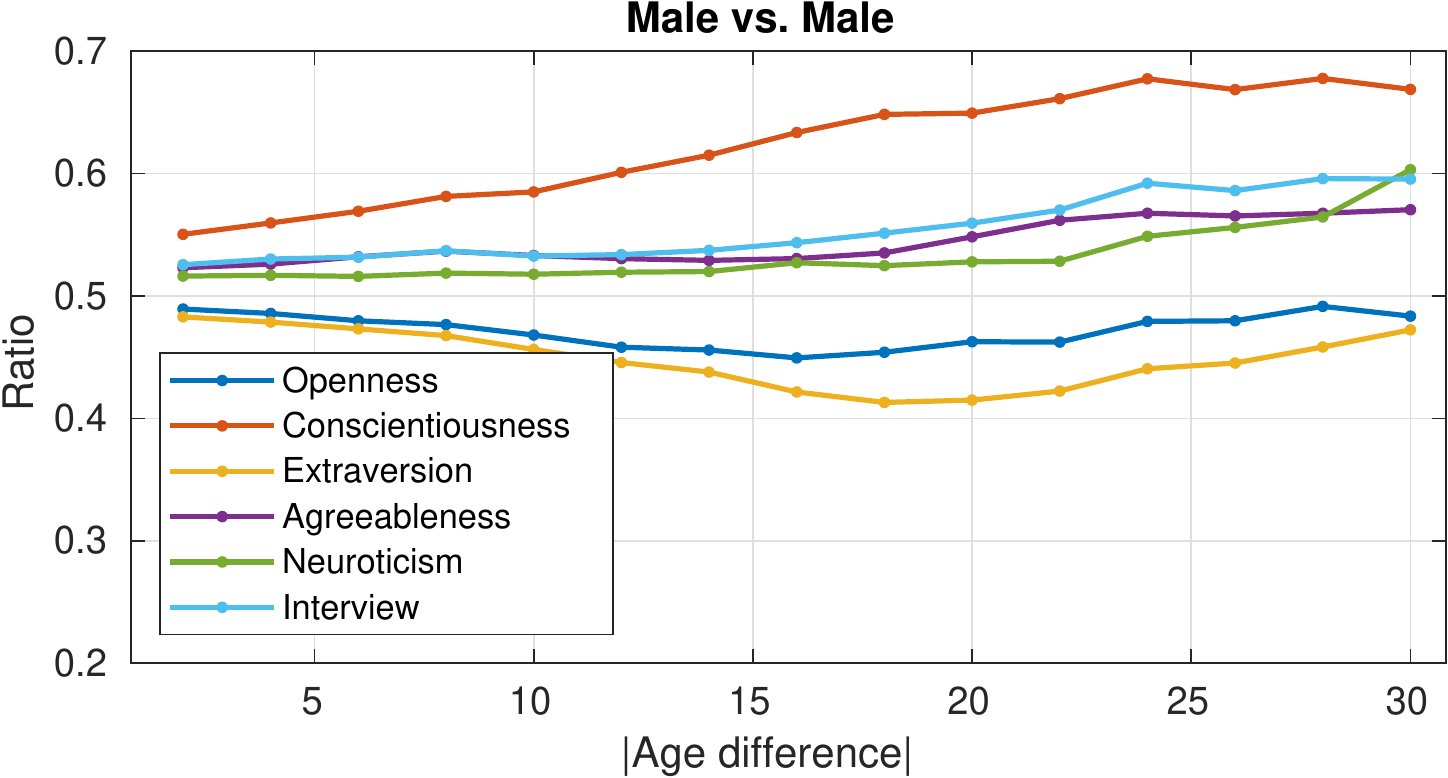}}
	\end{subfigure}
	\begin{subfigure}
	{\includegraphics[width=0.46\linewidth]{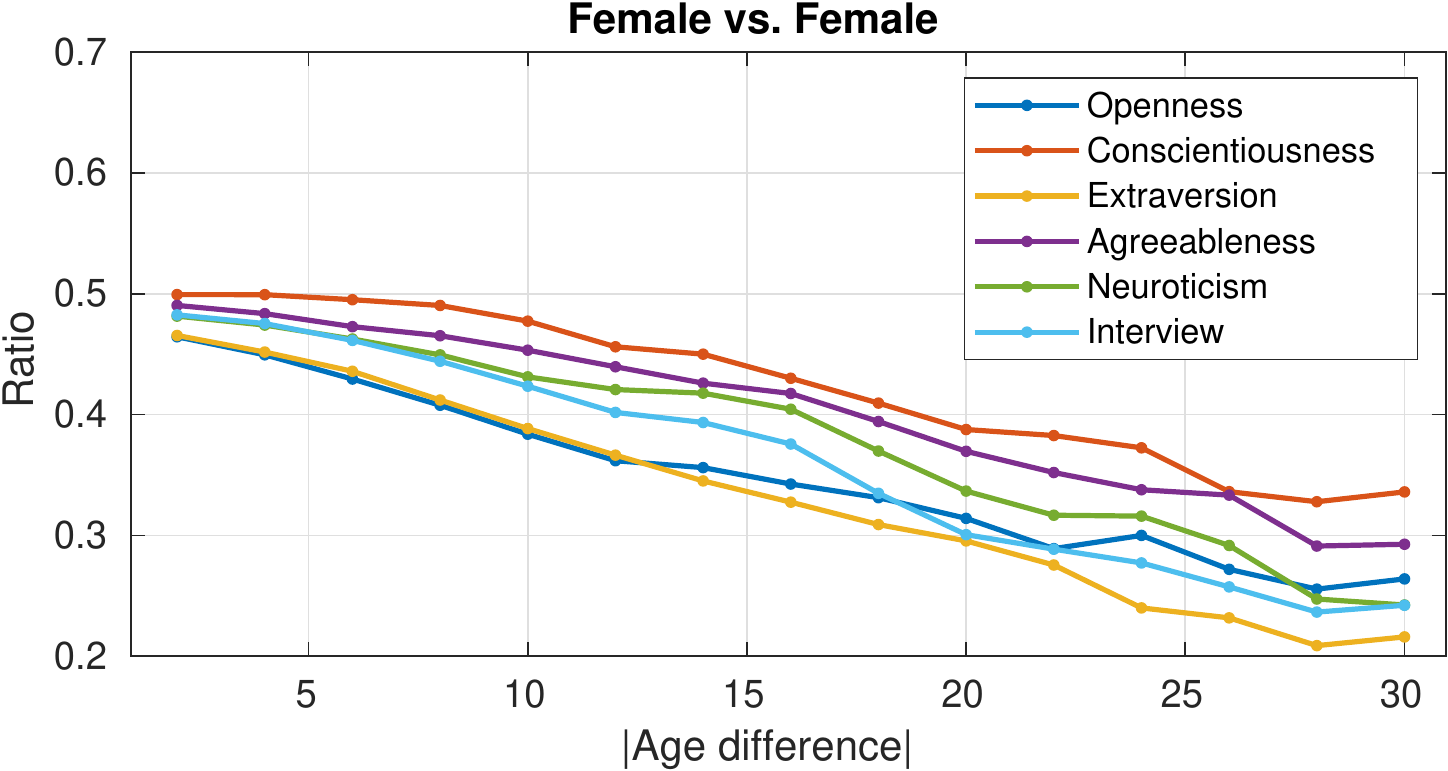}}
	\end{subfigure}
\caption{Perceived age bias (``Caucasian vs. Caucasian'' subset). Number of times an individual in a pair recognised as ``older'' was chosen, divided by the number of times a ``younger'' individual was selected, as a function of the perceived age difference between them. Ratio values $> 0.5$ indicates that ``older'' individuals are more frequently selected, while ratio $< 0.5$ the opposite, i.e., ``youngers'' are more frequently chosen. The plots show a overall bias towards older man and a clear bias towards younger women. Note, in this case, \textit{Neuroticism} relates to ``Emotion stability''.}	
	\label{fig:age_th_bias}
\end{figure*}

\subsubsection{Face attractiveness bias}\label{subsec:att}

Fig.~\ref{fig:att_th_bias} shows the number of times an individual in a pair recognised as ``more attractive'' was chosen, divided by the number of times an individual recognised as ``less attractive'' was selected, varying the face attractiveness difference between them. If face attractiveness had no influence, the ratio would be $0.5$, meaning that individuals recognised as more or less attractive were rated equally. Values higher than $0.5$ show a bias towards ``more attractive'' individuals, whereas values lower than $0.5$ a bias towards ``less attractive'' ones. As it can be seen, when face attractiveness difference between individuals increases, there is a higher fraction of individuals in the pairwise data recognised as ``more attractive'' being perceived as a more representative sample for a particular trait, suggesting that face attractiveness is biasing the annotations. This trend was observed to be stronger for some traits, and even stronger when both individuals being observed are women. Therefore, as face attractiveness difference between individuals increases, the number of pairs being analysed decreases (Fig.~\ref{fig:att_age_diff}, left image), which may affect the analysis, particularly for the cases where large differences are observed.

\subsubsection{Perceived age bias}\label{subsec:age}

Fig.~\ref{fig:age_th_bias} shows the number of times an individual in a pair perceived as ``older'' was chosen, divided the number of times an individual perceived as ``younger'' was selected, varying the perceived age difference between them. If perceived age had no influence, the ratio would be $0.5$, meaning that individuals recognised as older or younger were equally perceived. Values higher than $0.5$ show a bias towards older individuals, whereas values lower than $0.5$ a bias towards youngers. As it can be seen, annotators had an overall bias towards older men (except for traits ``O'' and ``E''), especially when age difference between individuals in a pair increases, and a bias towards younger women most of the time for all dimensions, suggesting that the perceived age attribute influenced data labelling in different ways. Therefore, as perceived age difference between individuals increases, the number of pairs being analysed decreases (see Fig.~\ref{fig:att_age_diff}, image on the right), which may affect the analysis, particularly for the cases where large differences are observed.

\section{Final Considerations}

This work used the First Impressions dataset as case study to expose how person perception can influence data labelling of a subjective task like personality. We analysed by the first time the original pairwise binary annotations provided with the FI dataset, and revealed the existence of different types of perception bias. This study also showed that the mechanism used to convert pairwise annotations to continuous values may magnify the biases if no special treatment is considered. Thus, this crucial step should be carefully revised, and possible negative consequences mitigated. In addition to gender and ethnicity biases, we analysed how the \textit{attractiveness halo effect} and the perception of age can affect data labelling of personality, derived from the pairwise annotations and face attributes automatically extracted. Although these perception biases have been widely studied in psychology and social sciences, the topic has received almost no attention from the computer vision community. 


After analysing the pairwise-based annotation setup of the FI dataset, our study suggests that new protocols need to pay more attention to the way the pairs are defined and presented to the annotators, since the pairs themselves can be a source of bias. Moreover, as perception is dependent on the observer, the analysis and correlation of attributes between annotators and people being annotated could explain how some biases are produced. However, this would require a dedicated discussion around privacy and ethical issues that goes beyond the scope of this work.



\section*{Acknowledgment}

This research was supported by Spanish projects TIN2015-66951-C2-2-R, RTI2018-095232-B-C22, and PID2019-105093GB-I00 (MINECO/FEDER, UE) and CERCA Programme/Generalitat de Catalunya. This work is partially supported by ICREA under the ICREA Academia programme.



{\small
\begin{thebibliography}{10}\itemsep=-1pt
	
	\bibitem{Agustsson2017}
	Eirikur Agustsson, Radu Timofte, Sergio Escalera, Xavier Baro, Isabelle Guyon,
	and Rasmus Rothe.
	\newblock Apparent and real age estimation in still images with deep residual
	regressors on appa-real database.
	\newblock In {\em International Conference on Automatic Face \& Gesture
		Recognition (FG)}, pages 87--94, 2017.
	
	\bibitem{Hendricks_2018_ECCV}
	Lisa Anne~Hendricks, Kaylee Burns, Kate Saenko, Trevor Darrell, and Anna
	Rohrbach.
	\newblock Women also snowboard: Overcoming bias in captioning models.
	\newblock In {\em The European Conference on Computer Vision (ECCV)}, September
	2018.
	
	\bibitem{Bird:2019:FML}
	Sarah Bird, Krishnaram Kenthapadi, Emre Kiciman, and Margaret Mitchell.
	\newblock Fairness-aware machine learning: Practical challenges and lessons
	learned.
	\newblock In {\em Int. Conference on Web Search and Data Mining}, pages
	834--835, 2019.
	
	\bibitem{breazeal2004social}
	Cynthia Breazeal.
	\newblock Social interactions in {HRI}: the robot view.
	\newblock {\em IEEE Transactions on Systems, Man, and Cybernetics, Part C
		(Applications and Reviews)}, 34(2):181--186, 2004.
	
	\bibitem{pmlr-v81-buolamwini18a}
	Joy Buolamwini and Timnit Gebru.
	\newblock Gender shades: Intersectional accuracy disparities in commercial
	gender classification.
	\newblock In {\em Conference on Fairness, Accountability and Transparency},
	volume~81, pages 77--91, 2018.
	
	\bibitem{Chen2016}
	Baiyu Chen, Sergio Escalera, Isabelle Guyon, V{\'i}ctor Ponce-L{\'o}pez, Nihar
	Shah, and Marc Oliu.
	\newblock Overcoming calibration problems in pattern labeling with pairwise
	ratings: Application to personality traits.
	\newblock In {\em European Conference on Computer Vision Workshop (ECCVW)},
	pages 419--432, 2016.
	
	\bibitem{chen2018my}
	Irene Chen, Fredrik~D Johansson, and David Sontag.
	\newblock Why is my classifier discriminatory?
	\newblock In {\em Advances in Neural Information Processing Systems}, pages
	3539--3550, 2018.
	
	\bibitem{Dekel:2020}
	Ron Dekel and Dov Sagi.
	\newblock Perceptual bias is reduced with longer reaction times during visual
	discrimination.
	\newblock {\em Communications Biology}, 3:59, 2020.
	
	\bibitem{Imagenet:2009}
	J. {Deng}, W. {Dong}, R. {Socher}, L. {Li}, {Kai Li}, and {Li Fei-Fei}.
	\newblock Imagenet: A large-scale hierarchical image database.
	\newblock In {\em Conference on Computer Vision and Pattern Recognition
		(CVPR)}, pages 248--255, 2009.
	
	\bibitem{Dimara:2020}
	E. {Dimara}, S. {Franconeri}, C. {Plaisant}, A. {Bezerianos}, and P.
	{Dragicevic}.
	\newblock A task-based taxonomy of cognitive biases for information
	visualization.
	\newblock {\em IEEE Transactions on Visualization and Computer Graphics},
	26(2):1413--1432, 2020.
	
	\bibitem{Escalante:IJCNN:2017}
	H.~J. {Escalante}, I. {Guyon}, S. {Escalera}, J.~C.~S. {Jacques Junior}, M.
	{Madadi}, X. {Baró}, S. {Ayache}, E. {Viegas}, Y. {Güçlütürk}, U.
	{Güçlü}, M.~A.~J. {van Gerven}, and R. {van Lier}.
	\newblock Design of an explainable machine learning challenge for video
	interviews.
	\newblock In {\em International Joint Conference on Neural Networks (IJCNN)},
	pages 3688--3695, 2017.
	
	\bibitem{Escalante:TAC:2020}
	H.~J. {Escalante}, H. {Kaya}, A. {Salah}, S. {Escalera}, Y. {Güç;lütürk},
	U. {Güçlü}, X. {Baró}, I. {Guyon}, J.~C.~S. {Jacques Junior}, M.
	{Madadi}, S. {Ayache}, E. {Viegas}, F. {Gurpinar}, A.~S. {Wicaksana}, C.
	{Liem}, M.~A.~J. {Van Gerven}, and R. {Van Lier}.
	\newblock Modeling, recognizing, and explaining apparent personality from
	videos.
	\newblock {\em IEEE Transactions on Affective Computing}, 2020.
	
	\bibitem{ethics:ai:eu}
	{European Commission}.
	\newblock Ethics guidelines for trustworthy ai.
	\newblock [online] Available at:
	\urlstyle{rm}\url{https://ec.europa.eu/digital-single-market/en/news/ethics-guidelines-trustworthy-ai}
	[Accessed 5 Oct. 2020], Apr 2019.
	
	\bibitem{friedler2019comparative}
	Sorelle~A Friedler, Carlos Scheidegger, Suresh Venkatasubramanian, Sonam
	Choudhary, Evan~P Hamilton, and Derek Roth.
	\newblock A comparative study of fairness-enhancing interventions in machine
	learning.
	\newblock In {\em Proceedings of the Conference on Fairness, Accountability,
		and Transparency}, pages 329--338. ACM, 2019.
	
	\bibitem{gu2019understanding}
	Jindong Gu and Daniela Oelke.
	\newblock Understanding bias in machine learning.
	\newblock {\em CoRR}, abs/1909.01866, 2019.
	
	\bibitem{Park:CVPR:2018}
	Dong Huk~Park, Lisa Anne~Hendricks, Zeynep Akata, Anna Rohrbach, Bernt Schiele,
	Trevor Darrell, and Marcus Rohrbach.
	\newblock Multimodal explanations: Justifying decisions and pointing to the
	evidence.
	\newblock In {\em IEEE Conference on Computer Vision and Pattern Recognition
		(CVPR)}, 2018.
	
	\bibitem{jacques:TAC:2019}
	Julio C.~S. {Jacques Junior}, Yagmur G{\"{u}}{\c{c}}l{\"{u}}t{\"{u}}rk, Marc
	P{\'{e}}rez, Umut G{\"{u}}{\c{c}}l{\"{u}}, Carlos And{\'{u}}jar, Xavier
	Bar{\'{o}}, Hugo~Jair Escalante, Isabelle Guyon, Marcel A.~J. van Gerven, Rob
	van Lier, and Sergio Escalera.
	\newblock First impressions: {A} survey on vision-based apparent personality
	trait analysis.
	\newblock {\em IEEE Transactions on Affective Computing (TAC)}, 2019.
	
	\bibitem{jacques:FG:2019}
	Julio C.~S. {Jacques Junior}, Cagri Ozcinar, Marina Marjanovic, Xavier
	Bar{\'{o}}, Gholamreza Anbarjafari, and Sergio Escalera.
	\newblock On the effect of age perception biases for real age regression.
	\newblock In {\em International Conference on Automatic Face \& Gesture
		Recognition (FG)}, pages 1--8, 2019.
	
	\bibitem{Jiang:2019}
	Heinrich Jiang and Ofir Nachum.
	\newblock Identifying and correcting label bias in machine learning.
	\newblock {\em CoRR}, abs/1901.04966, 2019.
	
	\bibitem{Joo:ICCV:2015}
	J. Joo, F.~F. Steen, and S.~C. Zhu.
	\newblock Automated facial trait judgment and election outcome prediction:
	Social dimensions of face.
	\newblock In {\em IEEE International Conference on Computer Vision (ICCV)},
	pages 3712--3720, 2015.
	
	\bibitem{LANGER2019231}
	Allison Langer, Ronit Feingold-Polak, Oliver Mueller, Philipp Kellmeyer, and
	Shelly Levy-Tzedek.
	\newblock Trust in socially assistive robots: Considerations for use in
	rehabilitation.
	\newblock {\em Neuroscience \& Biobehavioral Reviews}, 104:231 -- 239, 2019.
	
	\bibitem{liang2018scut}
	L. {Liang}, L. {Lin}, L. {Jin}, D. {Xie}, and M. {Li}.
	\newblock {SCUT-FBP5500}: A diverse benchmark dataset for multi-paradigm facial
	beauty prediction.
	\newblock In {\em International Conference on Pattern Recognition (ICPR)},
	pages 1598--1603, 2018.
	
	\bibitem{Lucker:1981}
	G.~William Lucker, William~E. Beane, and Robert~L. Helmreich.
	\newblock The strength of the halo effect in physical attractiveness research.
	\newblock {\em The Journal of Psychology}, 107(1):69--75, 1981.
	
	\bibitem{Matz:2017}
	S.~C. Matz, M. Kosinski, G. Nave, and D.~J. Stillwell.
	\newblock Psychological targeting as an effective approach to digital mass
	persuasion.
	\newblock {\em Proceedings of the National Academy of Sciences of the United
		States of America (PNAS)}, 114(48):12714--12719, 2017.
	
	\bibitem{todorov:2019:gender}
	DongWon Oh, Elinor~A. Buck, and Alexander Todorov.
	\newblock Revealing hidden gender biases in competence impressions of faces.
	\newblock {\em Psychological Science}, 30(1):65--79, 2019.
	
	\bibitem{Palmer:2016}
	Carl~L. Palmer and Rolfe~D. Peterson.
	\newblock Halo effects and the attractiveness premium in perceptions of
	political expertise.
	\newblock {\em American Politics Research}, 44(2):353--382, 2016.
	
	\bibitem{lopez2016chalearn}
	VP Ponce-Lopez, B Chen, A Places, M Oliu, C Corneanu, X Baro, HJ Escalante, I
	Guyon, and S Escalera.
	\newblock Cha{L}earn {LAP} 2016: First round challenge on first impressions -
	dataset and results.
	\newblock In {\em European Conference on Computer Vision Workshop (ECCVW)},
	pages 400--418, 2016.
	
	\bibitem{dario:TAC:2019}
	R.~D. {Pérez Principi}, C. {Palmero}, Julio C.~S. {Jacques Junior}, and S.
	{Escalera}.
	\newblock On the effect of observed subject biases in apparent personality
	analysis from audio-visual signals.
	\newblock {\em IEEE Transactions on Affective Computing}, pages 1--14, 2019.
	
	\bibitem{Quadrianto_2019_CVPR}
	Novi Quadrianto, Viktoriia Sharmanska, and Oliver Thomas.
	\newblock Discovering fair representations in the data domain.
	\newblock In {\em IEEE Conference on Computer Vision and Pattern Recognition
		(CVPR)}, 2019.
	
	\bibitem{Riva:2019}
	Silvia Riva, Ezekiel Chinyio, and Paul Hampton.
	\newblock Biased perceptions and personality traits attribution: Cognitive
	aspects in future interventions for organizations.
	\newblock {\em Frontiers in Psychology}, 9, 2019.
	
	\bibitem{Robinson_2020_CVPR_Workshops}
	Joseph~P. Robinson, Gennady Livitz, Yann Henon, Can Qin, Yun Fu, and Samson
	Timoner.
	\newblock Face recognition: Too bias, or not too bias?
	\newblock In {\em Conference on Computer Vision and Pattern Recognition (CVPR)
		Workshops}, 2020.
	
	\bibitem{Shen:2019:ACII}
	Judy~Hanwen Shen, Agata Lapedriza, and Rosalind~W. Picard.
	\newblock Unintentional affective priming during labeling may bias labels.
	\newblock In {\em International Conference on Affective Computing and
		Intelligent Interaction}, 2019.
	
	\bibitem{simonyan2014deep}
	Karen Simonyan and Andrew Zisserman.
	\newblock Very deep convolutional networks for large-scale image recognition.
	\newblock In {\em International Conference on Learning Representations (ICLR)},
	2015.
	
	\bibitem{Sixta:eccvw:2020}
	Tomáš Sixta, Julio C. S.~Jacques Junior, Pau Buch-Cardona, Eduard Vazquez,
	and Sergio Escalera.
	\newblock Fairface challenge at eccv 2020: Analyzing bias in face recognition.
	\newblock In {\em European Conference on Computer Vision Workshop (ECCVW)},
	2020.
	
	\bibitem{Talamas:2016}
	Sean~N. Talamas, Kenneth~I. Mavor, and David~I. Perrett.
	\newblock Blinded by beauty: Attractiveness bias and accurate perceptions of
	academic performance.
	\newblock {\em PLOS ONE}, 11(2):1--18, 2016.
	
	\bibitem{Timmerman:1980}
	Karl Timmerman and Jay Hewitt.
	\newblock Examining the halo effect of physical attractiveness.
	\newblock {\em Perceptual and Motor Skills}, 51(2):607--612, 1980.
	
	\bibitem{Todorov:2017:book}
	A. Todorov.
	\newblock {\em Face Value: The Irresistible Influence of First Impressions}.
	\newblock Princeton and Oxford: Princeton University Press, 2017.
	
	\bibitem{Torralba:cvpr:2011}
	A. {Torralba} and A.~A. {Efros}.
	\newblock Unbiased look at dataset bias.
	\newblock In {\em IEEE Conference on Computer Vision and Pattern Recognition
		(CVPR)}, pages 1521--1528, 2011.
	
	\bibitem{Vinciarelli:2013}
	Alessandro Vinciarelli and Gelareh Mohammadi.
	\newblock A survey of personality computing.
	\newblock {\em IEEE Transactions on Affective Computing (TAC)}, 5(3):273--291,
	2014.
	
	\bibitem{Wang_2019_ICCV}
	Tianlu Wang, Jieyu Zhao, Mark Yatskar, Kai-Wei Chang, and Vicente Ordonez.
	\newblock Balanced datasets are not enough: Estimating and mitigating gender
	bias in deep image representations.
	\newblock In {\em IEEE International Conference on Computer Vision (ICCV)},
	2019.
	
	\bibitem{Yan:ICMI:2020}
	Shen Yan, Di Huang, and Mohammad Soleymani.
	\newblock Mitigating biases in multimodal personality assessment.
	\newblock In {\em International Conference on Multimodal Interaction (ICMI)},
	page 361–369, 2020.
	
	\bibitem{Yucer_2020_CVPR_Workshops}
	Seyma Yucer, Samet Akcay, Noura Al-Moubayed, and Toby~P. Breckon.
	\newblock Exploring racial bias within face recognition via per-subject
	adversarially-enabled data augmentation.
	\newblock In {\em Conference on Computer Vision and Pattern Recognition (CVPR)
		Workshops}, 2020.
	
	\bibitem{Zebrowitz:2014}
	Leslie Zebrowitz and Robert Franklin.
	\newblock The attractiveness halo effect and the babyface stereotype in older
	and younger adults: Similarities, own-age accentuation, and older adult
	positivity effects.
	\newblock {\em Experimental aging research}, 40:375--393, 2014.
	
	\bibitem{75Zhang:SPL:2016}
	K. Zhang, Z. Zhang, Z. Li, and Y. Qiao.
	\newblock Joint face detection and alignment using multitask cascaded
	convolutional networks.
	\newblock {\em IEEE Signal Processing Letters}, 23(10):1499--1503, 2016.
	
	\bibitem{Zhao:2017}
	Jieyu Zhao, Tianlu Wang, Mark Yatskar, Vicente Ordonez, and Kai{-}Wei Chang.
	\newblock Men also like shopping: Reducing gender bias amplification using
	corpus-level constraints.
	\newblock {\em CoRR}, abs/1707.09457, 2017.
	
\end{thebibliography}
}
\end{document}